\documentclass[10pt,twocolumn,letterpaper]{article}

\usepackage{wacv}

\wacvfinalcopy

\usepackage{times}
\usepackage{epsfig}
\usepackage{graphicx}
\usepackage{amsmath}
\usepackage{amssymb}
\usepackage{booktabs}
\def\wacvPaperID{876} %

\wacvalgorithmstrack   %
\usepackage{algorithm}
\usepackage{algpseudocode}
\usepackage{multirow}
\usepackage[dvipsnames]{xcolor}
\usepackage{array}
\usepackage{booktabs}
\usepackage{enumitem}
\newcommand{\best}[1]{\textcolor{Blue}{\textbf{#1}}}
\newcommand{\second}[1]{\textcolor{Green}{\textit{#1}}}

\ifwacvfinal
\usepackage[breaklinks=true,bookmarks=false]{hyperref}
\else
\usepackage[pagebackref=true,breaklinks=true,colorlinks,bookmarks=false]{hyperref}
\fi

\ifwacvfinal
\fi

\begin{document}

\title{Hyperdimensional Feature Fusion for Out-of-Distribution Detection}

\author{Samuel Wilson\\
Queensland University of Technology\\
2 George St, Brisbane, QLD 4000, Australia\\
{\tt\small s84.wilson@hdr.qut.edu.au}
\and
Tobias Fischer\\
Queensland University of Technology\\
2 George St, Brisbane, QLD 4000, Australia\\
{\tt\small tobias.fischer@qut.edu.au}
\and
Niko Sünderhauf\\
Queensland University of Technology\\
2 George St, Brisbane, QLD 4000, Australia\\
{\tt\small niko.suenderhauf@qut.edu.au}
\and
Feras Dayoub\\
University of Adelaide\\
North Terrace, Adelaide, SA 5005, Australia\\
{\tt\small feras.dayoub@adelaide.edu.au}
}

\maketitle
\ifwacvfinal
\thispagestyle{empty}
\fi
\newcommand{\vect}[1]{\mathbf{ #1}}
\newcommand{\vectg}[1]{{\boldsymbol{ #1}}}
\newcommand{\ggo}{\ensuremath{\mathrm{g^2o}} }
\newcommand{\R}{\mathbb{R}}
\newcommand{\N}{\mathbb{N}}
\newcommand{\Z}{\mathbb{Z}}
\renewcommand{\P}{\mathbb{P}}
\newcommand{\tran}{^\top}
\newcommand{\T}{^\mathsf{T}}
\newcommand{\iT}{^{-\mathsf{T}}}
\newcommand{\inv}{^{-1}}
\newcommand{\func}[2]{\mathtt{#1}\left\{#2\right\}}
\newcommand{\sig}{\operatorname{sig}}
\newcommand{\diag}{\operatorname{diag}}
\newcommand{\argmin}{\operatornamewithlimits{argmin}}
\newcommand{\argmax}{\operatornamewithlimits{argmax}}
\newcommand{\RMSE}{\operatorname{RMSE}}
\newcommand{\RMSEpos}{\operatorname{RMSE}_\text{pos}}
\newcommand{\RMSEori}{\operatorname{RMSE}_\text{ori}}
\newcommand{\RPE}{\operatorname{RPE}}
\newcommand{\RPEpos}{\operatorname{RPE}_\text{pos}}
\newcommand{\RPEori}{\operatorname{RPE}_\text{ori}}
\newcommand{\rpe}{\varepsilon_{\vdelta}}
\newcommand{\achiError}{\bar{e}_{\chi^2}}
\newcommand{\chiError}{e_{\chi^2}}
\newcommand{\normal}[2]{\mathcal{N}\left(#1, #2\right)}
\newcommand{\uniform}[2]{\mathcal{U}\left(#1, #2\right)}
\newcommand{\pfrac}[2]{\frac{\partial #1}{\partial #2}}  %
\newcommand{\fracpd}[2]{\frac{\partial #1}{\partial #2}} %
\newcommand{\fracppd}[2]{\frac{\partial^2 #1}{\partial #2^2}}  %
\newcommand{\dd}{\mathrm{d}}  
\newcommand{\smd}[2]{\left\| #1 \right\|^2_{#2}}
\newcommand{\E}[1]{\text{\normalfont{E}}\left[ #1 \right]}     %
\newcommand{\Cov}[1]{\text{\normalfont{Cov}}\left[ #1 \right]} %
\newcommand{\Var}[1]{\text{\normalfont{Var}}\left[ #1 \right]} %
\newcommand{\Tr}[1]{\text{\normalfont{tr}}\left( #1 \right)}   %
\def\sgn{\mathop{\mathrm sgn}}    
\newcommand{\twovector}[2]{\begin{pmatrix} #1 \\ #2 \end{pmatrix}} %
\newcommand{\smalltwovector}[2]{\left(\begin{smallmatrix} #1 \\ #2 \end{smallmatrix}\right)} 
\newcommand{\threevector}[3]{\begin{pmatrix} #1 \\ #2 \\ #3 \end{pmatrix}} %
\newcommand{\fourvector}[4]{\begin{pmatrix} #1 \\ #2 \\ #3 \\ #4 \end{pmatrix}}  %
\newcommand{\smallthreevector}[3]{\left(\begin{smallmatrix} #1 \\ #2 \\ #3 \end{smallmatrix}\right)} %
\newcommand{\fourmatrix}[4]{\begin{pmatrix} #1 & #2 \\ #3 & #4 \end{pmatrix}} %
\newcommand{\vA}{\vect{A}}
\newcommand{\vB}{\vect{B}}
\newcommand{\vC}{\vect{C}}
\newcommand{\vD}{\vect{D}}
\newcommand{\vE}{\vect{E}}
\newcommand{\vF}{\vect{F}}
\newcommand{\vG}{\vect{G}}
\newcommand{\vH}{\vect{H}}
\newcommand{\vI}{\vect{I}}
\newcommand{\vJ}{\vect{J}}
\newcommand{\vK}{\vect{K}}
\newcommand{\vL}{\vect{L}}
\newcommand{\vM}{\vect{M}}
\newcommand{\vN}{\vect{N}}
\newcommand{\vO}{\vect{O}}
\newcommand{\vP}{\vect{P}}
\newcommand{\vQ}{\vect{Q}}
\newcommand{\vR}{\vect{R}}
\newcommand{\vS}{\vect{S}}
\newcommand{\vT}{\vect{T}}
\newcommand{\vU}{\vect{U}}
\newcommand{\vV}{\vect{V}}
\newcommand{\vW}{\vect{W}}
\newcommand{\vX}{\vect{X}}
\newcommand{\vY}{\vect{Y}}
\newcommand{\vZ}{\vect{Z}}
\newcommand{\va}{\vect{a}}
\newcommand{\vb}{\vect{b}}
\newcommand{\vc}{\vect{c}}
\newcommand{\vd}{\vect{d}}
\newcommand{\ve}{\vect{e}}
\newcommand{\vf}{\vect{f}}
\newcommand{\vg}{\vect{g}}
\newcommand{\vh}{\vect{h}}
\newcommand{\vi}{\vect{i}}
\newcommand{\vj}{\vect{j}}
\newcommand{\vk}{\vect{k}}
\newcommand{\vl}{\vect{l}}
\newcommand{\vm}{\vect{m}}
\newcommand{\vn}{\vect{n}}
\newcommand{\vo}{\vect{o}}
\newcommand{\vp}{\vect{p}}
\newcommand{\vq}{\vect{q}}
\newcommand{\vr}{\vect{r}}
\newcommand{\vt}{\vect{t}}
\newcommand{\vu}{\vect{u}}
\newcommand{\vv}{\vect{v}}
\newcommand{\vw}{\vect{w}}
\newcommand{\vx}{\vect{x}}
\newcommand{\vy}{\vect{y}}
\newcommand{\vz}{\vect{z}}
\newcommand{\valpha}{\vectg{\alpha}}
\newcommand{\vbeta}{\vectg{\beta}}
\newcommand{\vgamma}{\vectg{\gamma}}
\newcommand{\vdelta}{\vectg{\delta}}
\newcommand{\vepsilon}{\vectg{\epsilon}}
\newcommand{\vtau}{\vectg{\tau}}
\newcommand{\vmu}{\vectg{\mu}}
\newcommand{\vphi}{\vectg{\phi}}
\newcommand{\vPhi}{\vectg{\Phi}}
\newcommand{\vpi}{\vectg{\pi}}
\newcommand{\vPi}{\vectg{\Pi}}
\newcommand{\vPsi}{\vectg{\Psi}}
\newcommand{\vchi}{\vectg{\chi}}
\newcommand{\vvarphi}{\vectg{\varphi}}
\newcommand{\veta}{\vectg{\eta}}
\newcommand{\viota}{\vectg{\iota}}
\newcommand{\vkappa}{\vectg{\kappa}}
\newcommand{\vlambda}{\vectg{\lambda}}
\newcommand{\vLambda}{\vectg{\Lambda}}
\newcommand{\vnu}{\vectg{\nu}}
\newcommand{\vgo}{\vectg{\o}}
\newcommand{\vvarpi}{\vectg{\varpi}}
\newcommand{\vtheta}{\vectg{\theta}}
\newcommand{\vTheta}{\vectg{\Theta}}
\newcommand{\vvartheta}{\vectg{\vartheta}}
\newcommand{\vrho}{\vectg{\rho}}
\newcommand{\vsigma}{\vectg{\sigma}}
\newcommand{\vSigma}{\vectg{\Sigma}}
\newcommand{\vvarsigma}{\vectg{\varsigma}}
\newcommand{\vupsilon}{\vectg{\upsilon}}
\newcommand{\vomega}{\vectg{\omega}}
\newcommand{\vOmega}{\vectg{\Omega}}
\newcommand{\vxi}{\vectg{\xi}}
\newcommand{\vXi}{\vectg{\Xi}}
\newcommand{\vpsi}{\vectg{\psi}}
\newcommand{\vzeta}{\vectg{\zeta}}
\newcommand{\vzero}{\vect{0}}
\newcommand{\cA}{\mathcal{A}}
\newcommand{\cB}{\mathcal{B}}
\newcommand{\cC}{\mathcal{C}}
\newcommand{\cD}{\mathcal{D}}
\newcommand{\cE}{\mathcal{E}}
\newcommand{\cF}{\mathcal{F}}
\newcommand{\cG}{\mathcal{G}}
\newcommand{\cH}{\mathcal{H}}
\newcommand{\cI}{\mathcal{I}}
\newcommand{\cJ}{\mathcal{J}}
\newcommand{\cK}{\mathcal{K}}
\newcommand{\cL}{\mathcal{L}}
\newcommand{\cM}{\mathcal{M}}
\newcommand{\cN}{\mathcal{N}}
\newcommand{\cO}{\mathcal{O}}
\newcommand{\cP}{\mathcal{P}}
\newcommand{\cQ}{\mathcal{Q}}
\newcommand{\cR}{\mathcal{R}}
\newcommand{\cS}{\mathcal{S}}
\newcommand{\cT}{\mathcal{T}}
\newcommand{\cU}{\mathcal{U}}
\newcommand{\cV}{\mathcal{V}}
\newcommand{\cW}{\mathcal{W}}
\newcommand{\cX}{\mathcal{X}}
\newcommand{\cY}{\mathcal{Y}}
\newcommand{\cZ}{\mathcal{Z}}
\newcommand{\fA}{\mathfrak{A}}
\newcommand{\fB}{\mathfrak{B}}
\newcommand{\fC}{\mathfrak{C}}
\newcommand{\fD}{\mathfrak{D}}
\newcommand{\fE}{\mathfrak{E}}
\newcommand{\fF}{\mathfrak{F}}
\newcommand{\fG}{\mathfrak{G}}
\newcommand{\fH}{\mathfrak{H}}
\newcommand{\fI}{\mathfrak{I}}
\newcommand{\fJ}{\mathfrak{J}}
\newcommand{\fK}{\mathfrak{K}}
\newcommand{\fL}{\mathfrak{L}}
\newcommand{\fM}{\mathfrak{M}}
\newcommand{\fN}{\mathfrak{N}}
\newcommand{\fO}{\mathfrak{O}}
\newcommand{\fP}{\mathfrak{P}}
\newcommand{\fQ}{\mathfrak{Q}}
\newcommand{\fR}{\mathfrak{R}}
\newcommand{\fS}{\mathfrak{S}}
\newcommand{\fT}{\mathfrak{T}}
\newcommand{\fU}{\mathfrak{U}}
\newcommand{\fV}{\mathfrak{V}}
\newcommand{\fW}{\mathfrak{W}}
\newcommand{\fX}{\mathfrak{X}}
\newcommand{\fY}{\mathfrak{Y}}
\newcommand{\fZ}{\mathfrak{Z}}

\begin{abstract}
     We introduce powerful ideas from Hyperdimensional Computing into the challenging field of Out-of-Distribution (OOD) detection. In contrast to most existing works that perform OOD detection based on only a single layer of a neural network, we use similarity-preserving semi-orthogonal projection matrices to project the feature maps from \emph{multiple} layers into a common vector space. By repeatedly applying the \emph{bundling} operation $\oplus$, we create expressive class-specific descriptor vectors for all in-distribution classes. At test time, a simple and efficient cosine similarity calculation between descriptor vectors consistently identifies OOD samples with competitive performance to the current state-of-the-art whilst being significantly faster. We show that our method is orthogonal to recent state-of-the-art OOD detectors and can be combined with them to further improve upon the performance.
\end{abstract}
\section{Introduction} \label{sec:intro}
  Deep Neural Networks can achieve excellent performance on many vision tasks when the distribution of training data closely matches the test data. However, this assumption is often violated during deployment: especially in embodied AI application domains such as robotics and autonomous systems, objects and scenes that were not part of the training data distribution will inevitably be encountered. When confronted with such out-of-distribution (OOD) samples, deep neural networks tend to fail silently, producing overconfident but erroneous predictions~\cite{Hein_2019_CVPR,DBLP:journals/corr/abs-1808-03305} that can pose severe risks~\cite{sunderhauf2018limits}. It is therefore critically important that OOD samples are identified effectively during deployment.%
     \begin{figure}[t]
        \centering
        \includegraphics[width=1.0\columnwidth]{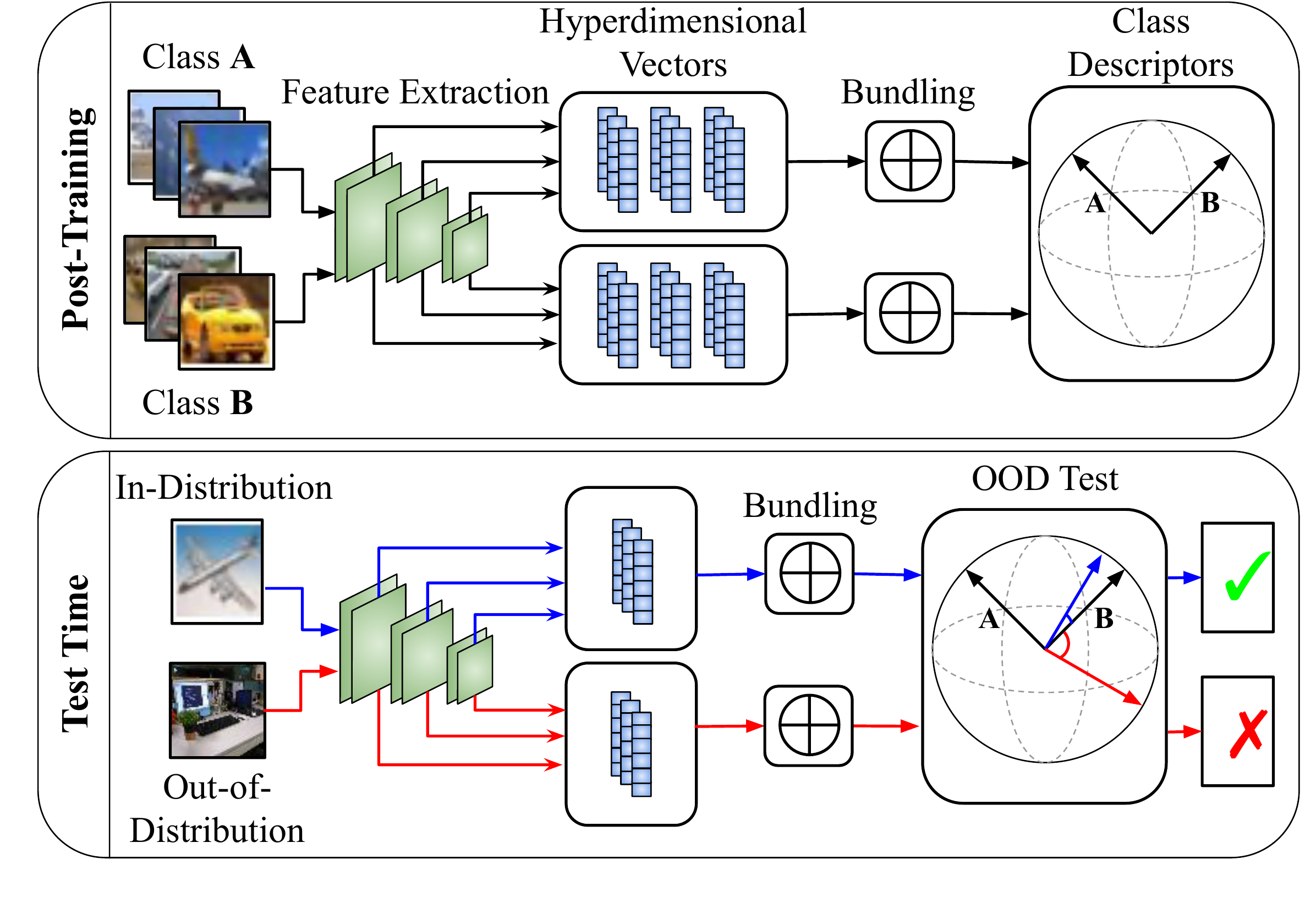}
        \caption{Illustration of our HDFF OOD detector. \textbf{Top:} After training, we extract the feature maps of in-distribution classes from multiple layers of the DNN and project them into a new hyperdimensional space with a similarity-preserving projection matrix. Using the bundling operation $\oplus$ we create class-specific hyperdimensional descriptor vectors. \textbf{Bottom:} During testing, we repeat feature extraction, projection and bundling to obtain a hyperdimensional image descriptor. OOD samples will produce descriptors with a large angular distance to all class descriptors (red vector).}
        \label{fig:the_hook}
    \end{figure}

    Previous techniques for OOD detection used softmax probabilities~\cite{hendrycks17baseline,liang2018enhancing} or distances in logit space~\cite{hendrycks2019anomalyseg} to distinguish in-distribution (ID) from OOD samples, making the implicit assumption that the features of a \emph{single} layer in a Deep Neural Network (DNN) contain sufficient information to identify OOD data. 
    However, deeper layers of a DNN can be more sensitive to OOD samples than the softmax probabilities~\cite{DBLP:journals/corr/abs-2012-03082}, which are often overconfident in the presence of OOD samples~\cite{Hein_2019_CVPR}.
    Recent work investigating the utility of multi-scale deep features from a DNN found that modelling ID classes from raw features is difficult without first reducing the number of dimensions~\cite{DBLP:journals/corr/abs-2012-03082,NEURIPS2018_abdeb6f5,DBLP:journals/corr/abs-1803-04765} due to the curse of dimensionality~\cite{10.1002/sam.11161}. 
    \par
    In this paper, we introduce the powerful concepts of similarity-preserving projection, binding and bundling from Hyper Dimensional Computing (HDC) \cite{SMOLENSKY1990159} to OOD detection. Figure \ref{fig:the_hook} provides a high-level description of our proposed OOD detector. We project the feature maps from \emph{multiple} layers of a network into a common high-dimensional vector space, using similarity-preserving semi-orthogonal projection matrices. This allows us to effectively \emph{fuse} the information contained in multiple layers: a series of \emph{bundling} operations $\oplus$ creates a class-specific high-dimensional vector representation for each of the in-distribution classes from the training dataset. During testing, we use the same principles of projection and bundling to obtain a single vector representation for a new input image. OOD samples can then be efficiently identified through cosine similarity operations with the class-specific representations.
    \par
    Our Hyperdimensional Feature Fusion (HDFF) approach relies on the pseudo-random nature of its projection matrices and the similarity-preserving properties of the bundling operation $\oplus$ to effectively fuse information contained in the feature maps across multiple layers in a deep network. HDFF avoids the difficulties of estimating densities in high-dimensional spaces \cite{DBLP:journals/corr/abs-2012-03082,NEURIPS2018_abdeb6f5}, removes the limitation of relying on a single layer \cite{hendrycks17baseline,hendrycks2019anomalyseg}, does not require to select informative layers based on example OOD data \cite{huang2021feature,Yu_2019_ICCV,liu2020energy}, and does not necessarily rely on expensive sampling techniques (such as ensembles \cite{NIPS2017_9ef2ed4b} or Monte Carlo Dropout \cite{Gal2016Dropout}) to identify OOD samples. HDFF can be applied to a trained network and does not require re-training or fine-tuning. Our HDFF detector competes with or exceeds state-of-the-art performance for OOD detection in conjunction with networks trained with standard cross-entropy loss or more sophisticated logit space shaping approaches~\cite{Zaeemzadeh_2021_CVPR}. 
    \par
    \noindent In summary, our contributions are the following:
    \begin{enumerate}[nosep]
        \item We propose a novel OOD detector based on the principles of Hyperdimensional Computing that is competitive with state-of-the-art OOD detectors whilst requiring significantly less training and inference time.
        \item We show that fusing features from multiple layers is critical for general OOD detection performance, as different layers vary in sensitivity to different OOD data, but fusion results in OOD performance that is approximately on par or even better than the best single layer.
        \item We show that the angles between HDFF vectors can be interpreted as visual similarity between inputs allowing for failure diagnosis.
    \end{enumerate}
    We release a publicly available code repository to replicate our experiments at: [released upon publication].

\section{Related Work} \label{sec:Background}
    Out-of-distribution (OOD) detection has been addressed in the context of image classification \cite{liang2018enhancing,liu2020energy,Zaeemzadeh_2021_CVPR,hendrycks17baseline} and semantic segmentation \cite{hendrycks2019anomalyseg,Di_Biase_2021_CVPR,10.1007/978-3-030-33676-9_3,Blum2021} with a diverse set of solutions to each of the problems. In this section, we highlight overarching concepts and most relevant contributions in three areas: i) Many approaches try to extrapolate the properties of the distribution of OOD data by utilising a small subset of the data to fine-tune on. ii) When this information is unavailable (OOD agnostic), distances or densities in the softmax probabilities and logit spaces are used. iii) Comparatively few bodies of work exist that use multiple feature maps for OOD detection. %
    
    \textbf{OOD Fine-tuning} 
    Some prior work~\cite{DBLP:journals/corr/abs-2106-09022,huang2021feature,Yu_2019_ICCV,liu2020energy,liang2018enhancing,Di_Biase_2021_CVPR,10.1007/978-3-030-33676-9_3} requires the availability of an OOD dataset to train or fine-tune the network or adjust the OOD detection process. However, such approaches are problematic. By definition, the OOD dataset has to always be incomplete: during deployment, samples that do not match the assumptions embedded in the OOD dataset can still be encountered, leaving such methods exposed to high OOD risk. In contrast, we do not require prior knowledge about the characteristics of OOD data, instead we directly model the in-distribution data and measure deviations from this set.
    
    \textbf{OOD Agnostic} Most approaches that do not rely on the availability of OOD data during training or validation analyse the model outputs to identify OOD data during deployment. Early work~\cite{hendrycks17baseline} showed the maximum of a model's softmax output to be a good baseline. Expanding on this idea, \cite{hendrycks2019anomalyseg} proposed using the maximum of the unnormalised output logits. Additional methods focus on extracting more information from the outputs \cite{Oberdiek2020} or learning calibrated confidence estimates \cite{devries2018learning,lee2018training}. Model averaging techniques such as Deep Ensembles \cite{NIPS2017_9ef2ed4b}, Monte Carlo (MC) Dropout~\cite{Gal2016Dropout,hendrycks2019anomalyseg} and SWA-G~\cite{NEURIPS2019_118921ef,Zaeemzadeh_2021_CVPR} often improve the OOD detection performance. Training networks with specialised losses that impose a beneficial structure on the feature space has been growing in popularity~\cite{miller2021class,Zaeemzadeh_2021_CVPR,miller2021gmmdet}. E.g.~the recent 1D Subspaces~\cite{Zaeemzadeh_2021_CVPR} OOD estimator encourages the network to learn orthogonal features in the layer before the logits. The approaches discussed above attempt OOD detection based on only a \emph{single} layer in the network. In contrast, we show that utilising \emph{multiple} layers is beneficial for OOD detection.
    
    \textbf{Multi-Feature OOD Detection} Newer approaches incorporate and analyse multi-scale feature maps extracted from a DNN to detect OOD samples. The use of feature maps and additional information extracted from a network has been shown to set a new state-of-the-art in both OOD segmentation \cite{Di_Biase_2021_CVPR,10.1007/978-3-030-33676-9_3} and classification \cite{Dong_2022_CVPR} when used as input to an auxiliary OOD detection network. The requirement on an auxiliary network necessarily entails training on OOD, or proxy-OOD, data that is strongly representative of the OOD samples likely to be drawn from target distribution. Related work with auxiliary networks in online performance monitoring demonstrates that this class of networks can be robust to failures of the primary network without prior knowledge of the exact failures that are likely to be encountered \cite{DBLP:journals/corr/abs-2011-07750,8968525}. Avoiding these problems altogether, recent works have tried to model in-distribution classes without an auxiliary network \cite{DBLP:journals/corr/abs-2012-03082,NEURIPS2018_abdeb6f5,DBLP:journals/corr/abs-1803-04765}. However, the curse of dimensionality requires these methods to rely on PCA reductions \cite{DBLP:journals/corr/abs-2012-03082,NEURIPS2018_abdeb6f5} or random projections \cite{DBLP:journals/corr/abs-1803-04765}, losing information from the feature maps. %
    By contrast, we use techniques from Hyperdimensional Computing to model our in-distribution data without the need for an auxiliary network. We seek to build upon evidence in the literature \cite{DBLP:journals/corr/abs-2012-03082,pmlr-v119-sastry20a,Lin_2021_CVPR, Dong_2022_CVPR} that multi-scale features benefit OOD detection performance.

\section{Hyperdimensional Computing} \label{sec:VSA}
    Hyperdimensional Computing (HDC), also known as Vector Symbolic Architectures, is the field of computation with vectors in hyperdimensional, very large, spaces. HDC techniques leverage the large amount of expressive power in the hyperdimensional (HD) space to model associations between data points using redundant HD vector encodings. Redundant means that in spaces of $10^4$ dimensionality it can be seen that two random vectors are almost guaranteed to be within 5 degrees of orthogonal \cite{schlegel2020comparison} known as \emph{quasi-orthogonality}. The associations between data points are represented as vectors, and the set of the representative vectors is known as the \emph{associative memory} \cite{9107175}. Construction of the associative memory is done through the structured combination of HD vectors using a set of standard operations. For our application we are primarily interested in the \emph{bundling $\oplus$} and \emph{binding $\otimes$} operation for combining feature vectors, and the \textit{encoding} operation for projecting low dimensional representations into our target HD space.
    
    \textbf{Bundling} The bundling $\oplus$ operation is used to store a representation of multiple input vectors that retains similarity to all of the input vectors \cite{Plate97}. Concretely, given random vectors $\va$, $\vb$ and $\vc$, the vector $\va$ will be similar to the bundles $\va\oplus\vb$, $\va\oplus\vc$ and $\va\oplus\vb\oplus\vc$, although in the final case it will be less similar as the bundled vector needs to be similar to all 3 input vectors. Typically, the bundling operation is implemented as element-wise addition with some form of normalisation to the required space for the architecture \cite{schlegel2020comparison}. Normalisation steps that are commonly seen are normalisation to a magnitude of one \cite{Gosmann2019,Gallant2013} or cutting/thresholding to a range of desired values \cite{Gayler1998MultiplicativeBR}. Without normalisation, the bundling operation is commutative and associative; when normalisation steps are added the associative aspect of the operation is approximate, $(\va\oplus\vb)\oplus\vc \approx \va\oplus(\vb\oplus\vc)$ \cite{schlegel2020comparison}. 
    
    \textbf{Binding} The binding $\otimes$ operation is used to combine a set of vectors into one representation that is dissimilar to all of the input vectors \cite{Plate97}. Given the quasi-orthogonality property of HD spaces, this entails that in all likelihood, the binding operation will generate a vector orthogonal to all of the input vectors in the cosine similarity space. The second core trait of the binding operation is that it approximately preserves the similarity of two vectors before and after binding if they are bound to the same target vector. More precisely, for a set of vectors in the same hyperdimensional space $\va$, $\vb$ and $\vc$, that $sim(\va, \vb) \approx sim(\va\otimes \vc, \vb\otimes \vc)$.
    
    \textbf{Encoding} The format of the original data will not always be in the hyperdimensional space that is required. Encoding addresses this by projecting each data point from the original space into the new HD space. The selection of data encoding into a HD space depends on the type of the original data \cite{10.1145/2934583.2934624,10.1145/3453688.3461749}. An important principle of encoding is that distances in the input are preserved in the output \cite{10.1145/3453688.3461749}, much like the binding operation. Examples of encoding include multiplication with a projection matrix \cite{9388914}, fractional binding~\cite{fracbind} that preserves real numbered differences, or the encoding proposed in~\cite{10.1145/3453688.3461749} that preserves similarities of vectors over time and spatial coordinates.
    \par
    We provided only a brief introduction to HDC, for more in-depth discussions and comparisons between known architectures we encourage the reader to consider \cite{schlegel2020comparison,Neubert2019,Kanerva2009}.
\section{Hyperdimensional Feature Fusion} \label{sec:HDFF}
We propose Hyperdimensional Feature Fusion (HDFF), a novel OOD detection method that applies the HDC concepts of Encoding and Bundling to the features from \emph{multiple} layers in a deep neural network, without requiring re-training or any prior knowledge of the OOD data. Our core idea is to project the feature maps from multiple layers into a common vector space, using similarity-preserving semi-orthogonal projection matrices. Through a series of bundling operations, we create a class-specific vector-shaped representation for each of the classes in the training dataset. During deployment, we repeat the projection and bundling steps for a new input image, and use the cosine similarity to the class representatives to identify OOD samples. We provide pseudo code to assist with describing our HDFF detector in Algorithm \ref{alg:demo}.

\textbf{Preliminaries}
From a pretrained network $f$, we can extract feature maps $\vm_l$ from different layers $l\in \{1,\dots,L\}$ in the the network. We apply a pooling operation across the spatial dimensions to reduce the tensor-shaped (height $\times$ width $\times$ channels) feature maps to vector representations $\vv_l$ of length (channels) for each layer; we ablate the effects of different pooling in the Supplementary Material. Since the feature maps of different layers have a different number of channels $c_l$, we need to project the vectors $\vv_l$ into a common $m$-dimensional vector space $\R^m$ in order to combine them. Conventional sizes for HD spaces range around $10^3$ - $10^4$ \cite{Neubert_2021,9388914,10.1145/3453688.3461749}, meaning that typically our HD space will be much larger than our original space $m>>c$.

\textbf{Feature Encoding}
Any $m\times c$ matrix $\vP$ projects a $c$-dimensional vector $\vv$ into a $m$-dimensional space by left-multiplying: $\vP\vv$. However, preserving the cosine similarity of the projected vectors is a crucial consideration for our application in OOD detection. We therefore impose that the projection matrices $\vP_l$ preserve the inner product of any two vectors $\va$ and $\vb$ in their original and their projected vector spaces. Formally, this means that 
\begin{eqnarray}
(\vP\cdot\va)\T \cdot (\vP\cdot\vb) &=& \va\T \cdot \vb \\
\va\T\cdot \vP\T \cdot \vP \cdot \vb &=& \va\T \cdot \vb
\end{eqnarray}
Above condition is satisfied by any matrix that fulfils the requirement $\vP\T\vP = \vI$, which is the defining property of a semi-orthogonal matrix.
Using the approach of~\cite{DBLP:journals/corr/SaxeMG13}, we therefore create a unique, pseudo-random, semi-orthogonal projection matrix $\vP_l$ for each of the considered layers $l$. These project the feature vectors $\vv_l$ into a common $m$-dimensional vector space:
\begin{equation}
    \vh_l = \vP_l\vv_l \quad \text{so that} \;\;\vh_l \in \R^m \;\; \forall l \in \{1,\dots, L\}    
\end{equation}

\begin{algorithm}[t]
    \begin{algorithmic}
        \item \textbf{Inputs:} Set of images from the training set $x^{\{1,\dots,I\}}$ with corresponding labels $j^{\{1,\dots,I\}}$. Set of models in an ensemble $f_{\{1,\dots,E\}}$ that produces a set of feature maps $\vm_{\{1,\dots,L\}}$ given an input $x^{(i)}$.
        \item \textbf{Outputs:} Set of class descriptor vectors per input model $d^{(e)}_{\{1,\dots,c\}}$ and a set of ensemble class descriptor vectors $d^{*}$
        \For{e $\in$ \{1,\dots,E\}}
            \For{i $\in$ \{1,\dots,I\}}
            \State $m^{(i)}_{\{1,\dots,L\}} \gets f_{e}(x^{(i)})$
                \For{l $\in$ \{1,\dots,L\}}
                \State $v^{(i)}_{l} \gets Pool(m^{(i)}_{l})$
                \State $h^{(i)}_{l} \gets P_{l} \cdot v^{(i)}_l$
                \EndFor
            \State $y^{(i)} \gets h^{(i)}_{1} \oplus h^{(i)}_{2} \oplus ... \oplus h^{(i)}_{L}$
            \State $d^{(e)}_{j^{(i)}} \gets d^{(e)}_{j^{(i)}} \oplus y^{(i)}$
            \EndFor
        \EndFor
        \State $d^{*} \gets (d^{(1)} \otimes z^{(1)}) \oplus ... \oplus (d^{(E)} \otimes z^{(E)}) $
        \caption{Computation of the model-wise class descriptors and ensemble descriptor vectors from training set.}\label{alg:demo}
    \end{algorithmic}
\end{algorithm}

\textbf{Feature Bundling}
Following the previous steps, we obtain a set of $L$ high-dimensional vectors $\vh_l^{(i)}$ for an input image $\vx^{(i)}$. Since all $\vh_l^{(i)}$ are elements of the same vector space, we can use the bundling operation $\oplus$ to combine them into a single vector $\vy^{(i)}$ that serves as an expressive descriptor for the input image $\vx^{(i)}$:
\begin{equation}
    \vy^{(i)} = \bigoplus_{l=1}^L \vh_l^{(i)} = \bigoplus_{l=1}^L \vP_l \cdot \vv_l^{(i)}
    \label{eq:image_descriptor}
\end{equation}
As explained in Section~\ref{sec:VSA}, the resulting vector $\vy^{(i)}$ will be cosine-similar to all contributing vectors $\vh_l^{(i)}$, but dissimilar to vectors from $\R^m$ that were not part of the bundle. Essentially, $\vy^{(i)}$ provides a summary of the feature vectors $\vv_l^{(i)}$ of the entire network for a single image $\vx^{(i)}$.

By bundling the descriptors of all images from the training set belonging to class $c$, we obtain a class-specific descriptor $\vd_c$:
\begin{equation}
    \label{eq:hd}
    \vd_c = \bigoplus_{j \in\mathbb{I}_c} \vy^{(j)}
\end{equation}
where $\mathbb{I}_c$ denotes the set of indices of the training images belonging to class $c$. 

As discussed in Section~\ref{sec:VSA}, the bundling operation $\oplus$ can be implemented in various ways. We implement $\oplus$ to be an element-wise sum, without truncation. %

\textbf{Out-of-Distribution Detection}
During testing or deployment, an image $\vx$ can be identified as OOD by obtaining its image descriptor $\vy$ according to~(\ref{eq:image_descriptor}), and calculating the cosine similarity to each of the class-specific descriptors $\vd_c$. Let $\theta$ be the angle to the class descriptor $\vd_c$ that is most similar to $\vy$
\begin{equation}
    \label{eq:theta}
    \theta = \min_{c}\cos^{-1}\left(\frac{\vy\T\vd_c}{\|\vy\| \cdot \|\vd_c\|}\right) ,\; c\in \{1,\dots,C\},
\end{equation}
The input $\vx$ is then treated as OOD if $\theta$ is bigger than a threshold:  $\theta > \theta^*$.

\textbf{Ensembling}
While HDFF does not rely on ensembling, we briefly show that our method is amenable to ensembling to further boost performance (however at the cost of added compute). When using a set of pretrained networks $f_{e}$ in an ensemble $e \in \{1,\dots,E\}$ we collect inputs from all models and fuse them into singular image and class descriptors $\vy_{*}$ and $\vd^{*}$ respectively. For each model $f_{\{1,\dots,E\}}$ the same process using equations~(\ref{eq:image_descriptor}) and~(\ref{eq:hd}) is used to compute the set of class descriptors for each model in the ensemble $\vd^{\{1,\dots,E\}}$. To ensure that each class descriptor is sufficiently distinct from all other descriptors, a set of random hyperdimensional vectors $\vz^{(e)}$ are generated and bound $\otimes$ to the class descriptor. By bundling the bound class descriptors, we obtain the ensemble class descriptor $\vd^{*}$:
\begin{equation}
    \vd^{*} =  \bigoplus_{e=1}^E \vd^{(e)} \otimes \vz^{(e)}
    \label{eq:ensemble_descriptor}
\end{equation}
When new input samples arrive $\vy_{*}$ is computed according to~(\ref{eq:ensemble_descriptor}) by substituting $\vd$ for $\vy$. OOD detection is done according to~(\ref{eq:theta}) using $\vd^{*}$ and $\vy_{*}$.

\section{Experiments} \label{sec:experiments}
    We conduct a series of experiments to demonstrate the efficacy of Hyperdimensional Feature Fusion for OOD detection. We compare HDFF to the current state-of-the-art in the typical far-OOD settings, where the distributions of the of ID and OOD are very dissimilar (e.g.~CIFAR10 $\rightarrow$ SVHN), and the more challenging near-OOD where the ID and OOD datasets are drawn from similar distributions (e.g.~CIFAR10 $\rightarrow$ CIFAR100). Further, we report the results of some critical ablation studies, in particular, we identify which layers are most sensitive to OOD data and how sensitive HDFF is to the decision parameter $\theta^*$ during deployment.
    
    \subsection{Experimental Setup}
    \textbf{Datasets} 
    For our comparisons to existing OOD detectors we use a wide array of popular datasets composed from multiple recent state-of-the-art OOD detectors. We construct our evaluation suite as the combination of datasets from~\cite{liang2018enhancing,pmlr-v119-sastry20a,Lin_2021_CVPR}. For our comparison to existing OOD detectors, we use the popular dataset splits for near- and far-OOD detection using CIFAR10 and CIFAR100~\cite{Krizhevsky_2009_17719} as the ID datasets. For our ID sets, we use the 50,000 training examples for training and computation of the class bundles, whilst the 10,000 testing images are used as our unseen ID data. For the near-OOD configurations, the test set of whichever CIFAR dataset is not being used for training will be used as the OOD set. For the far-OOD detection settings, we use a suite of benchmarks: iSUN, TinyImageNet~\cite{deng2009imagenet} (cropped and resized: TINc and TINr),  LSUN~\cite{yu15lsun} (cropped and resized: LSUNc and LSUNr)\footnote{Download links for OOD datasets can be found in the following repository: https://github.com/facebookresearch/odin}, SVHN~\cite{SVHN}, MNIST~\cite{deng2012mnist}, KMNIST~\cite{DBLP:journals/corr/abs-1812-01718}, FashionMNIST~\cite{xiao2017/online}, and Textures~\cite{cimpoi14describing}. In the interest of brevity, we only show the settings where CIFAR10 is the ID set, the settings of CIFAR100 as ID are provided in the Supplementary Material.
    
    \textbf{Evaluation Metrics}
    We consider the standard metrics~\cite{Zaeemzadeh_2021_CVPR,pmlr-v119-sastry20a} for our comparison to existing OOD literature. \textbf{AUROC:} The Area Under The Receiver Operating Characteristic curve corresponds to the area under the ROC curve with true positive rate (TPR) on the y-axis and false positive rate (FPR) on the x-axis. AUROC can be interpreted as the probability that an OOD sample will be given a higher score than an ID sample \cite{hendrycks2019anomalyseg}. \textbf{FPR95:} The FPR95 metric reports the FPR at a critical threshold which achieves a minimum of 95\% in TPR. \textbf{Detection Error:} Detection Error indicates the minimum misclassification probability with respect to the critical threshold. \textbf{F1:} F1-score corresponds to the harmonic mean of precision and recall. We make use of F1-score to evaluate the general binarisation performance in our ablations. In the interests of brevity, we only evaluate using the AUROC metric for our evaluations. We provide additional evaluations with FPR95 and Detection Error in the Supplementary Material.
    
    \textbf{Baselines}
    We divide our selected baselines and SOTA comparisons into two evaluation streams, these being; \textbf{statistics} and \textbf{training}. The \textbf{statistics} stream contains methods that are generally applied post-hoc to a pretrained network, requiring no training and only minimal calibration to represent in-distribution data. The \textbf{training} stream contains methods that allow for the retraining of the base network or the training of an auxiliary monitoring network. The distinction between the two streams is important for a fair comparison as the methods in the \textbf{training} stream have a significantly longer calibration (training) time as well as biases towards the OOD data, either explicitly through proxy-OOD training or implicitly through assumptions about the formation of OOD data with a custom loss.
    
    In the \textbf{statistics} stream we compare against: maximum softmax probability~\cite{hendrycks17baseline} (MSP), max logit~\cite{hendrycks2019anomalyseg} (ML), gramian matrices~\cite{pmlr-v119-sastry20a} (Gram) and energy-based model~\cite{liu2020energy} without calibration on OOD data. For the \textbf{training} stream we compare against: NMD~\cite{Dong_2022_CVPR}, Spectral Discrepancy trained with the 1D Subspaces methodology~\cite{Zaeemzadeh_2021_CVPR} (1DS), Deep Deterministic Uncertainty~\cite{mukhoti2022deep} (DDU)  and MOOD~\cite{Lin_2021_CVPR}.
    
    \textbf{Implementation} 
    We follow the evaluation procedure defined in~\cite{Zaeemzadeh_2021_CVPR,liang2018enhancing}, implementing the standard WideResNet~\cite{DBLP:journals/corr/ZagoruykoK16} network with a depth of 28 and a width of 10. In the \textbf{statistics} stream, this model is trained to convergence with standard cross-entropy loss. Custom loss objectives are restricted to the \textbf{training} stream. 
    
    As HDFF is a post-hoc statistics-based method, it is innately orthogonal to the methods contained within the \textbf{training} stream and can be combined with many of them in a post-hoc fashion. To demonstrate the robustness of HDFF we combine it with two of the other state-of-the-art detectors, 1D Subspaces~\cite{Zaeemzadeh_2021_CVPR} and NMD~\cite{Dong_2022_CVPR}. The \textit{HDFF-1DS} method applies as described in Section \ref{sec:HDFF} but the base model has been trained with the 1D Subspaces custom objective. The second model \textit{HDFF-MLP} trains an auxiliary MLP as a binary ID/OOD classifier using the generated HDFF image descriptor vectors $\vy^{(i)}$ of perturbed (proxy-OOD) and normal (ID) images from the ID training set as input into the MLP alike NMD~\cite{Dong_2022_CVPR}. 
   
    We re-implement the MSP, ML and Gram detectors for the WideResNet architecture, for all other methods, excluding NMD~\cite{Dong_2022_CVPR} and MOOD~\cite{Lin_2021_CVPR}, we report the published results on the same architecture. At the time of writing, a publicly available implementation of NMD~\cite{Dong_2022_CVPR} is unavailable and as such we utilise their published results in the zero-shot OOD scenario on ResNet34~\cite{He_2016_CVPR}. MOOD~\cite{Lin_2021_CVPR} is built upon a custom network architecture that enables early exits during inference, we report the published results on this custom architecture due to the unclear applicability of the method to other network architectures.
    
    When applying HDFF and Gram \cite{pmlr-v119-sastry20a} to the WideResNet model we attempt to faithfully recreate the hook locations of Gram from the original architecture. Specifically, features are recorded from the outputs of almost all of the following modules within the network: Conv2d, ReLU, BasicBlock, NetworkBlock and shortcut connections. In total, there are 76 features extracted per sample and hence the same number of semi-orthogonal projection matrices $\vP$ are generated for HDFF.
    
    Unless otherwise stated, we operate in a hyperdimensional space of $10^4$ dimensions, we ablate this hyperparameter in the Supplementary Material. Before projecting feature maps into the hyperdimensional space as per (\ref{eq:hd}), we apply mean-centering by subtracting the layer-wise mean activations (obtained from the training set) from all $\vm_l$. For pooling we apply max pooling over the spatial dimensions to reduce our feature maps $\vm_l$ to a vector representation, we ablate the effect of this choice in the Supplementary Material.
 
\begin{table}[t]
  \centering
  \scriptsize
    \begin{tabular}{l|cccccc}
     \multicolumn{7}{c}{Statistics Stream - CIFAR10} \\ \hline
     \multicolumn{1}{c|}{OOD} & HDFF & HDFF-Ens & Gram & MSP & ML & Energy \\
     \multicolumn{1}{c|}{Dataset} & (Ours) & (Ours) & \cite{pmlr-v119-sastry20a} & \cite{hendrycks17baseline} & \cite{hendrycks2019anomalyseg} & \cite{liu2020energy} \\ \hline
       iSun  & 99.2 & \second{99.3} & \best{99.9} & 96.4 & 97.8 & 92.6 \\
       TINc  & 98.3 & \second{98.4} & \best{99.4} & 95.4 & 96.8 & - \\
       TINr  & 99.2 & \second{99.4} & \best{99.8} & 95.0 & 96.5 & - \\
       LSUNc & 96.2 & 96.8 & \second{98.1} & 95.7 & 97.1 & \best{98.4} \\
       LSUNr & 99.2 & \second{99.4} & \best{99.9} & 96.5 & 98.0 & 94.2 \\
       SVHN  & \second{99.4} & \best{99.5} & \second{99.4} & 96.0 & 97.2 & 91.0 \\
       MNIST & 99.6 & \second{99.7} & \best{99.97} & 89.4 & 90.6 & - \\
       KMNIST & 99.0 & \second{99.1} & \best{99.98} & 92.7 & 93.4 & - \\
       FMNIST & 98.7 & \second{99.1} & \best{99.8} & 93.6 & 95.2 & - \\
       Textures   & 94.5 & \second{94.8} & \best{98.2} & 92.7 & 93.5 & 85.2 \\
       CIFAR100 & 75.4 & 75.8 & 79.4 & \best{87.8} & \second{87.3} & - \\ \hline
    \multicolumn{1}{c|}{\textbf{Average}}
          & 96.2 & \second{96.5} & \best{97.6} & 93.7 & 94.9 & 93.2  \\
    \end{tabular}\vspace*{0.1cm}%
\caption{OOD detection results for the against the methods contained belonging to the \textbf{statistics} stream. Comparison metric is AUROC, higher is better. \best{Best} results are shown in \best{blue and bold}, \second{second} best results are shown in \second{green and italics}. The ensemble in \textit{HDFF-Ens} always consists of 5 models. HDFF and Gram are consistently the top two performers across the significant majority of the far-OOD detection settings.}
  \label{tab:eval:stat}%
\end{table}%

\begin{table}[t]
\setlength{\tabcolsep}{4pt}
  \centering
  \scriptsize
    \begin{tabular}{l|cccccc}
    \multicolumn{7}{c}{Training Stream - CIFAR10} \\ \hline
     \multicolumn{1}{c|}{OOD} & HDFF-MLP & HDFF-1DS & 1DS & NMD & DDU & MOOD \\
     \multicolumn{1}{c|}{Dataset} & (Ours) & (Ours) & \cite{Zaeemzadeh_2021_CVPR} & \cite{Dong_2022_CVPR} & \cite{mukhoti2022deep} & \cite{Lin_2021_CVPR} \\ \hline
       iSun  & \best{99.99} & \second{99.9} & - & \second{99.9} & - & 93.0 \\
       TINc  & \best{99.9} & \second{99.7} & 98.1 & 99.2* & 91.1* & - \\
       TINr  & \best{99.96} & \second{99.8} & 98.5 & - & 91.1* & - \\
       LSUNc & 98.2 & 99.1 & \best{99.4} & 98.8 & - & \second{99.2} \\
       LSUNr & \best{99.99} & \second{99.9} & 99.3 & - & - & 93.3 \\
       SVHN  & 84.8 & \second{99.2} & - & \best{99.6} & 97.9 & 96.5 \\
       MNIST & \second{99.4} & 99.3 & - & - & - & \best{99.8} \\
       KMNIST & 98.6 & \second{99.3} & - & - & - & \best{99.9} \\
       FMNIST & \best{99.6} & \second{99.3} & - & - & - & -  \\
       Textures   &  \second{97.4} & 97.3 & - & \best{98.9} & - & 93.3\\
       CIFAR100 & 69.9 & \second{90.7} & - & 90.1 & \best{91.3} & - \\ \hline
    \textbf{Average}
          & 95.2 & \second{98.5} & \best{98.8} & 97.8 & 94.6 & 95.0  \\
    \end{tabular}\vspace*{0.1cm}%
\caption{OOD detection results for the against the methods contained belonging to the \textbf{training} stream. Comparison metric is AUROC, higher is better. \best{Best} results are shown in \best{blue and bold}, \second{second} best results are shown in \second{green and italics}. Published results that are unclear which variant of TIN they correspond to are identified with a *. The incorporation of HDFF into pre-existing pipelines leads to consistently improving or comparable results, demonstrating the robust nature of HDFF.}
  \label{tab:eval:train}%
\end{table}%

\begin{figure}[t]
    \centering
    \includegraphics[width=1.0\columnwidth]{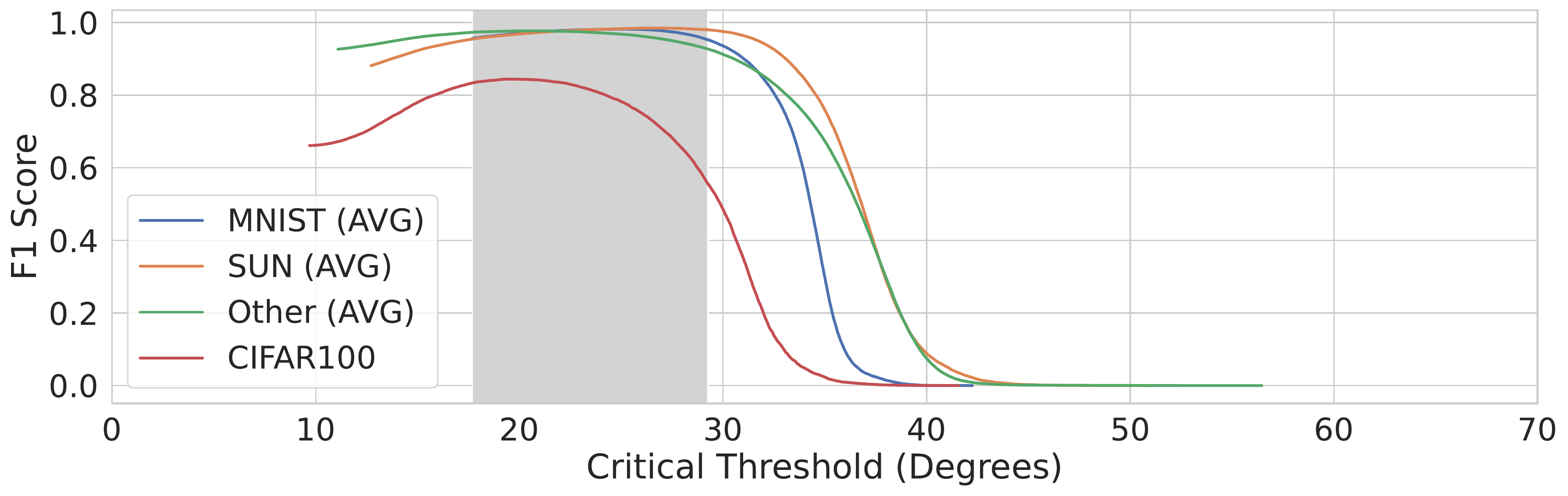}
    \vspace{0.2cm}
    \caption{F1-score for binarisation at different critical values of angular distance to closest class descriptor. The model used is the 1D Subspaces trained WideResNet. The grey region corresponds to binarisations that would produce a result within 5\% of the maximum F1-score achieved for all far-OOD datasets. To avoid clutter, far-OOD datasets have been grouped: i) MNIST (AVG) contains KMNIST, MNIST and FashionMNIST. ii) SUN (AVG) contains iSUN, LSUNr and LSUNc. iii) Other (AVG) contains all other far-OOD datasets. As expected, the near-OOD task (CIFAR100) leads to significantly lower thresholds compared to all far-OOD tasks.
    }
    \label{fig:criticalThresh}
\end{figure}

    \subsection{Results and Discussion} \label{sec:results}
    Table \ref{tab:eval:stat} compares the results of our HDFF OOD detector on the AUROC metric to all of the methods in the \textbf{statistics} stream under both the near- and far-OOD settings. In the far-OOD setting HDFF and Gram are consistently the top performers with small performance differences of less than 1\% AUROC between the two methods on most OOD dataset configurations. This finding indicates that the feature representations of the in-distribution data from both Gram and HDFF are powerful for the far-OOD detection task. On this note, we identify that the vector representation of HDFF is far more compact than the square matrix representations from Gram. The difference in representation complexity accounts for the performance differences, but also introduces large gaps in computational performance with gram requiring $\approx$4.5x longer per inference pass compared to HDFF as described in Section \ref{sec:complexity}. 
    
    We note that the MSP detector outperforms all other methods in the \textbf{statistics} stream in the near-OOD detection setting (CIFAR10 as in-distribution and CIFAR100 as OOD). Considering that HDFF is effectively detecting deviations in convolutional feature activations this would indicate that images with similar features are being grouped with in-distribution classes, this behaviour is discussed more in Section \ref{sec:interpret}.

    Table \ref{tab:eval:train} compares the results of the methods in the \textbf{training} stream on the AUROC metric in the near- and far-OOD settings. The first finding from this table is the broadly powerful nature of the HDFF vector representation in combination with the MLP. Across the board, the majority of the top performing results are from the HDFF MLP detector demonstrating the power of the HDFF representation when combined with latest state-of-the-art detectors. 
    
    Secondly and more specifically, we note that when HDFF is applied to the 1D Subspaces trained model it improves upon the performance of the Spectral Discrepancy detector on 3 out of the 4 comparable benchmarks. We additionally note that the Spectral discrepancy detector requires 50 SWA-G~\cite{maddox_2019_simple} samples to achieve these performance levels whilst the HDFF detector requires only one inference pass, mandating a 5000\% increase in computational time when using Spectral Discrepancy. These two findings combined reinforce the claim that HDFF is generally applicable to a wide range of models and training regimes. 
    
    \subsection{Computational Efficiency} \label{sec:complexity}
    HDC techniques are commonly used for computation or learning in low-power situations \cite{9388914,10.1145/3453688.3461749} and as such, we expect HDFF to introduce minimal computational overhead. For a full pass of the CIFAR10 test set, HDFF takes 7.0$\pm$0.9s compared to a standard inference pass at 6.0$\pm$0.7s, averaged over 5 independent runs. We note that this $\approx$17\% increase is comparatively minor considering the large performance gains that HDFF boasts over the MSP and ML detectors. By comparison, the closest equivalent method to ours, Gram~\cite{pmlr-v119-sastry20a} takes 31.4$\pm$0.3s to complete a full inference pass over the CIFAR10 test set, resulting in an $\approx$4.5x longer inference time than HDFF. Additionally, we expect the computational efficiency of Gram to drop on larger networks due to the gram matrices scaling $O(n^2)$ in computation and memory requirements with the number of channels whereas HDFF scales linearly $O(n)$ with the number of channels. 
    
    Characteristic of belonging to the \textbf{statistics} stream, HDFF requires far less calibration or training than other methods belonging to the \textbf{training} stream. In particular, the closest comparative method, NMD, prescribes a training regime of 60-100 epochs for the MLP detector, whereas HDFF only requires a single epoch to collect the full in-distribution statistics, resulting in a minimum 60x decrease in computational time. Additionally, HDFF can be applied post-hoc to common networks, requiring no additional computation in the training or fine-tuning of the network. We further note that HDFF with only a single inference pass competes with or exceeds the Spectral Discrepancy detector~\cite{Zaeemzadeh_2021_CVPR} that requires 50 MC samples to be collected, necessitating a minimum inference increase of 5000\%.
    
    \subsection{Critical Threshold Ablation}
    During deployment, it is often more useful if an OOD detector produces a binarisation of ID vs OOD rather than a raw OOD estimate; the grey regions in Figure~\ref{fig:criticalThresh} show a range of critical thresholds that will produce performance reasonably close to optimal in this setting. Specifically, the grey region in Figure~\ref{fig:criticalThresh} shows a region of confidence where any critical value would produce an F1-score within 5\% of the maximum value achieved on each far-OOD dataset, an approximate standard of reasonable performance. The near-OOD detection task is plotted but does not contribute towards the definition of the grey region due to the severe differences between the near- and far-OOD tasks. As we can see, a large region of critical values around the range of 18-29 degrees will result in generally good performance for far-OOD detection.

    \subsection{Which Layers Are Most Sensitive To OOD Data?}
\begin{figure}[t]
            \centering
            \includegraphics[width=1.0\columnwidth]{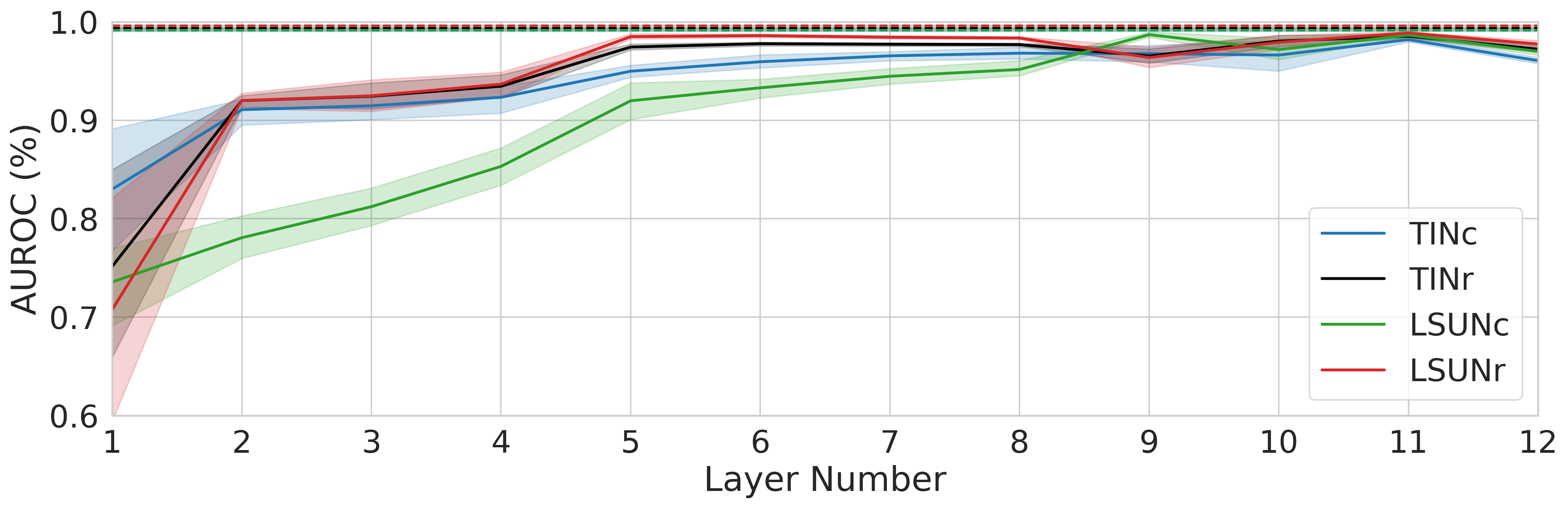}
            \caption{Comparison of effectiveness for OOD detection of each layer individually with the AUROC measure. CIFAR10 is the in-distirbution dataset. Mean and 95\% confidence interval over 5 randomly initialised models trained with the 1D Subspaces methodology. Dotted lines correspond to the results attained by the fusion of all 12 BasicBlocks. Performance of individual layers shows a trend toward later layers being more effective at detecting OOD data. The fusion of feature maps from across the network provides good general performance across all datasets and does not require calibration on OOD data to set.}
            \label{fig:LayerAblate}
        \end{figure}

        Congruent with other multi-layer OOD detectors \cite{Dong_2022_CVPR, pmlr-v119-sastry20a} we ablate the sensitivity of individual layers with respect to OOD samples from different target distributions. Figure \ref{fig:LayerAblate} demonstrates the effectiveness of individual layers when they are used in isolation for OOD detection based on our HDFF OOD estimator. For the sake of readability, we explicitly change the hook function to only collect features from the outputs of the 12 BasicBlocks and only consider the TIN and LSUN datasets for simplicity. We see in both ID settings that earlier features in the network appear to be less reliable at detecting OOD samples. This trend is particularly obvious in the CIFAR10 setting where there is a clear upwards trend as the layer number increases. This observation lends weight towards the suggestions in prior work that shallow layers in a DNN are unable to or at least are less effective at detecting OOD data \cite{DBLP:journals/corr/abs-2012-03082}.
        \par
        The dotted lines, and related 95\% confidence shaded area, for each dataset correspond to the fusion of information from all layers; we consider these lines the best that can be reasonably achieved within the bounds of the OOD task. In Figure \ref{fig:LayerAblate} we observe that no single individual layer is able to detect the OOD data as well as the fusion of all feature maps.
        Critically, we note two points in favour of the fusion of features: (1) without OOD data the optimal layer combination for a given data set is unknown, and (2) the optimal layer combination is not consistent between OOD data configurations as shown by the drop in performance on LSUNc. To summarise, note that if there is no access to OOD data at training time to determine which layer(s) are the best then the fusion of feature maps from across the network often provides the best performance or a close approximation. We provide extra ablations against other metrics and the CIFAR100 ID set in the Supplementary Material.
    
    \subsection{Distance of Features as Visual Similarity} \label{sec:interpret}
    As HDFF uses similarity preserving projections, the angular distance directly represents the differences between two input sets of raw features. Since these features are extracted from a deep CNN we can consider the angular distance between any two vectors to be a proxy for their visual dissimilarity. This definition leads to intuitive understandings of how HDFF behaves and identifying failure cases; we discuss these points here.
   
\begin{figure}[t]
        \centering
        \includegraphics[width=1\columnwidth]{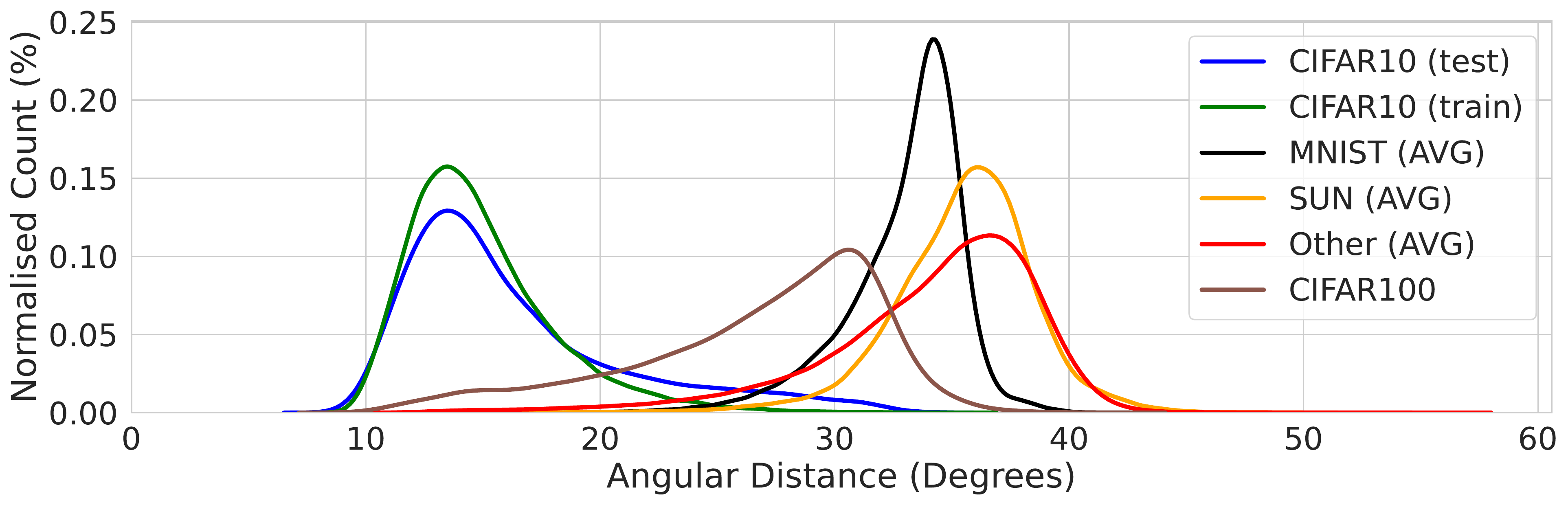}
        \caption{KDE estimate of separation between ID and OOD sets based on minimum angular distance to closest class representative in the CIFAR10 setting. To avoid clutter, far-OOD datasets have been grouped: i) MNIST (AVG) contains KMNIST, MNIST and FashionMNIST. ii) SUN (AVG) contains iSUN, LSUNr and LSUNc. iii) Other (AVG) contains all other far-OOD datasets. Overlap between the test and OOD distributions can be considered erroneous samples.}
        \label{fig:CosDist}
    \end{figure}

    Using our definition, since HDFF separates based on visual similarity, we can infer that failures in OOD detection are due either to ID samples being visually dissimilar to the training set or OOD samples are as similar, if not more so, to the training set than the ID samples. To aid in understanding this, Figure \ref{fig:CosDist} visualises the differences in distributions of angular distance between the ID and OOD datasets in the CIFAR10 ID setting through a KDE estimate (for smoothing) over binned angular distances on the HDFF 1D Subspaces model. The area of overlap between the test set and any OOD set can be considered as erroneous detection. Inspecting Figure \ref{fig:CosDist}, we observe that, in the far-OOD settings, errors due to dissimilar ID samples are more likely to appear due to the distributional shift between the training and testing distributions, i.e.~false positives. By contrast, in the near-OOD detection task we see that a significant number of errors are due to OOD samples appearing very similar to ID samples, i.e.~false negatives.
    
\begin{figure}[t]
        \centering
        \includegraphics[width=1\columnwidth]{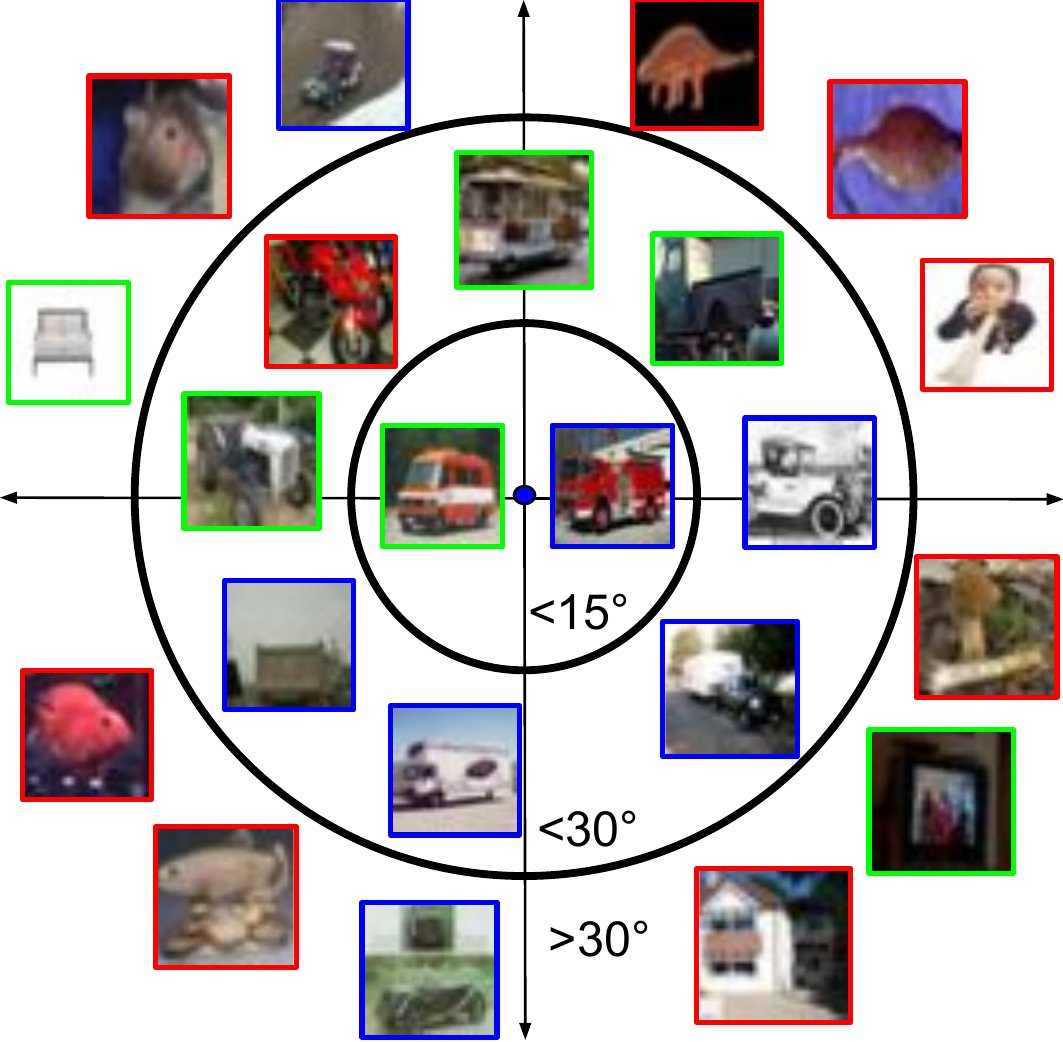}
        \caption{Sample images from in-distribution (\textcolor{blue}{CIFAR10, blue}), near-OOD (\textcolor{green}{CIFAR100, green}) and far-OOD (\textcolor{red}{TINc, red}) datasets with their approximate distance to the nearest class descriptor vector, corresponding to the CIFAR10 ID truck class. The underlying model is the 1D Subspaces trained WideResNet. Distances to the class bundle (centre blue dot) can be approximately inferred from which circle the sample is encapsulated by. This figure reinforces the hypothesis that HDFF separates based on visual similarity to the in-distribution class as truck-like objects appear within the inner circle, vehicle-like objects in the middle region and heavily dissimilar object falling outside both of those regions.
        }
        \label{fig:sim_axis}
        \vspace{-5pt}
    \end{figure}

    As a more concrete example, Figure \ref{fig:sim_axis} demonstrates HDFF separating input samples based on the angle to the nearest class descriptor vector, in this case, the CIFAR10 truck class. Consistent with our previous assertions, we observe in Figure \ref{fig:sim_axis} that samples that are $< 15^{\circ}$ from the class descriptor vector appear visually very similar, with no far-OOD samples occupying this range. In the range of $15-30^{\circ}$ we observe that samples from all datasets still have vehicle-like appearance, but whether or not these accurately represent a truck is debatable; this region is still predominantly populated by ID and near-OOD samples. Once outside the $30^{\circ}$ ring, we see that the significant majority of samples do not appear vehicle-like with the very few ID samples in this region having the truck visually obscured; this region is dominated by far-OOD samples.

\section{Conclusion} \label{sec:conclusion}
This paper introduced powerful ideas from hyperdimensional computing into the important task of OOD detection. We investigate the sensitivity of individual layers to OOD data and find that the fusion of feature maps provides the best general performance with no requirements for OOD data to fine-tune on. We perform competitively with state-of-the-art OOD detection methods with the added benefit of significantly reducing the computational costs associated with the current state-of-the-art. We show the interpretation of cosine distance as a proxy for visual similarity allows for additional failure diagnosis capabilities over competing methods. In this paper, we utilised the simple but powerful element-wise addition for bundling, however, this is one of potentially many applications of HDC to DNNs, opening new future research directions.

\ifwacvfinal
\noindent\textbf{Acknowledgements:} The authors acknowledge continued support from the Queensland University of Technology (QUT) through the Centre for Robotics. TF was partially supported by funding from ARC Laureate Fellowship FL210100156 and Intel’s Neuromorphic Computing Lab. We also thank Peer Neubert from Chemnitz University of Technology for greatly insightful discussions on the topic of Hyperdimensional Computing.
\fi

{\small
\bibliographystyle{ieee_fullname}
\bibliography{refs}
}

\end{document}


\title{Hyperdimensional Feature Fusion for Out-of-Distribution Detection - Supplementary Material}

\maketitle
\ifwacvfinal
\thispagestyle{empty}
\fi
\newcommand{\vect}[1]{\mathbf{ #1}}
\newcommand{\vectg}[1]{{\boldsymbol{ #1}}}
\newcommand{\ggo}{\ensuremath{\mathrm{g^2o}} }
\newcommand{\R}{\mathbb{R}}
\newcommand{\N}{\mathbb{N}}
\newcommand{\Z}{\mathbb{Z}}
\renewcommand{\P}{\mathbb{P}}
\newcommand{\tran}{^\top}
\newcommand{\T}{^\mathsf{T}}
\newcommand{\iT}{^{-\mathsf{T}}}
\newcommand{\inv}{^{-1}}
\newcommand{\func}[2]{\mathtt{#1}\left\{#2\right\}}
\newcommand{\sig}{\operatorname{sig}}
\newcommand{\diag}{\operatorname{diag}}
\newcommand{\argmin}{\operatornamewithlimits{argmin}}
\newcommand{\argmax}{\operatornamewithlimits{argmax}}
\newcommand{\RMSE}{\operatorname{RMSE}}
\newcommand{\RMSEpos}{\operatorname{RMSE}_\text{pos}}
\newcommand{\RMSEori}{\operatorname{RMSE}_\text{ori}}
\newcommand{\RPE}{\operatorname{RPE}}
\newcommand{\RPEpos}{\operatorname{RPE}_\text{pos}}
\newcommand{\RPEori}{\operatorname{RPE}_\text{ori}}
\newcommand{\rpe}{\varepsilon_{\vdelta}}
\newcommand{\achiError}{\bar{e}_{\chi^2}}
\newcommand{\chiError}{e_{\chi^2}}
\newcommand{\normal}[2]{\mathcal{N}\left(#1, #2\right)}
\newcommand{\uniform}[2]{\mathcal{U}\left(#1, #2\right)}
\newcommand{\pfrac}[2]{\frac{\partial #1}{\partial #2}}  %
\newcommand{\fracpd}[2]{\frac{\partial #1}{\partial #2}} %
\newcommand{\fracppd}[2]{\frac{\partial^2 #1}{\partial #2^2}}  %
\newcommand{\dd}{\mathrm{d}}  
\newcommand{\smd}[2]{\left\| #1 \right\|^2_{#2}}
\newcommand{\E}[1]{\text{\normalfont{E}}\left[ #1 \right]}     %
\newcommand{\Cov}[1]{\text{\normalfont{Cov}}\left[ #1 \right]} %
\newcommand{\Var}[1]{\text{\normalfont{Var}}\left[ #1 \right]} %
\newcommand{\Tr}[1]{\text{\normalfont{tr}}\left( #1 \right)}   %
\def\sgn{\mathop{\mathrm sgn}}    
\newcommand{\twovector}[2]{\begin{pmatrix} #1 \\ #2 \end{pmatrix}} %
\newcommand{\smalltwovector}[2]{\left(\begin{smallmatrix} #1 \\ #2 \end{smallmatrix}\right)} 
\newcommand{\threevector}[3]{\begin{pmatrix} #1 \\ #2 \\ #3 \end{pmatrix}} %
\newcommand{\fourvector}[4]{\begin{pmatrix} #1 \\ #2 \\ #3 \\ #4 \end{pmatrix}}  %
\newcommand{\smallthreevector}[3]{\left(\begin{smallmatrix} #1 \\ #2 \\ #3 \end{smallmatrix}\right)} %
\newcommand{\fourmatrix}[4]{\begin{pmatrix} #1 & #2 \\ #3 & #4 \end{pmatrix}} %
\newcommand{\vA}{\vect{A}}
\newcommand{\vB}{\vect{B}}
\newcommand{\vC}{\vect{C}}
\newcommand{\vD}{\vect{D}}
\newcommand{\vE}{\vect{E}}
\newcommand{\vF}{\vect{F}}
\newcommand{\vG}{\vect{G}}
\newcommand{\vH}{\vect{H}}
\newcommand{\vI}{\vect{I}}
\newcommand{\vJ}{\vect{J}}
\newcommand{\vK}{\vect{K}}
\newcommand{\vL}{\vect{L}}
\newcommand{\vM}{\vect{M}}
\newcommand{\vN}{\vect{N}}
\newcommand{\vO}{\vect{O}}
\newcommand{\vP}{\vect{P}}
\newcommand{\vQ}{\vect{Q}}
\newcommand{\vR}{\vect{R}}
\newcommand{\vS}{\vect{S}}
\newcommand{\vT}{\vect{T}}
\newcommand{\vU}{\vect{U}}
\newcommand{\vV}{\vect{V}}
\newcommand{\vW}{\vect{W}}
\newcommand{\vX}{\vect{X}}
\newcommand{\vY}{\vect{Y}}
\newcommand{\vZ}{\vect{Z}}
\newcommand{\va}{\vect{a}}
\newcommand{\vb}{\vect{b}}
\newcommand{\vc}{\vect{c}}
\newcommand{\vd}{\vect{d}}
\newcommand{\ve}{\vect{e}}
\newcommand{\vf}{\vect{f}}
\newcommand{\vg}{\vect{g}}
\newcommand{\vh}{\vect{h}}
\newcommand{\vi}{\vect{i}}
\newcommand{\vj}{\vect{j}}
\newcommand{\vk}{\vect{k}}
\newcommand{\vl}{\vect{l}}
\newcommand{\vm}{\vect{m}}
\newcommand{\vn}{\vect{n}}
\newcommand{\vo}{\vect{o}}
\newcommand{\vp}{\vect{p}}
\newcommand{\vq}{\vect{q}}
\newcommand{\vr}{\vect{r}}
\newcommand{\vt}{\vect{t}}
\newcommand{\vu}{\vect{u}}
\newcommand{\vv}{\vect{v}}
\newcommand{\vw}{\vect{w}}
\newcommand{\vx}{\vect{x}}
\newcommand{\vy}{\vect{y}}
\newcommand{\vz}{\vect{z}}
\newcommand{\valpha}{\vectg{\alpha}}
\newcommand{\vbeta}{\vectg{\beta}}
\newcommand{\vgamma}{\vectg{\gamma}}
\newcommand{\vdelta}{\vectg{\delta}}
\newcommand{\vepsilon}{\vectg{\epsilon}}
\newcommand{\vtau}{\vectg{\tau}}
\newcommand{\vmu}{\vectg{\mu}}
\newcommand{\vphi}{\vectg{\phi}}
\newcommand{\vPhi}{\vectg{\Phi}}
\newcommand{\vpi}{\vectg{\pi}}
\newcommand{\vPi}{\vectg{\Pi}}
\newcommand{\vPsi}{\vectg{\Psi}}
\newcommand{\vchi}{\vectg{\chi}}
\newcommand{\vvarphi}{\vectg{\varphi}}
\newcommand{\veta}{\vectg{\eta}}
\newcommand{\viota}{\vectg{\iota}}
\newcommand{\vkappa}{\vectg{\kappa}}
\newcommand{\vlambda}{\vectg{\lambda}}
\newcommand{\vLambda}{\vectg{\Lambda}}
\newcommand{\vnu}{\vectg{\nu}}
\newcommand{\vgo}{\vectg{\o}}
\newcommand{\vvarpi}{\vectg{\varpi}}
\newcommand{\vtheta}{\vectg{\theta}}
\newcommand{\vTheta}{\vectg{\Theta}}
\newcommand{\vvartheta}{\vectg{\vartheta}}
\newcommand{\vrho}{\vectg{\rho}}
\newcommand{\vsigma}{\vectg{\sigma}}
\newcommand{\vSigma}{\vectg{\Sigma}}
\newcommand{\vvarsigma}{\vectg{\varsigma}}
\newcommand{\vupsilon}{\vectg{\upsilon}}
\newcommand{\vomega}{\vectg{\omega}}
\newcommand{\vOmega}{\vectg{\Omega}}
\newcommand{\vxi}{\vectg{\xi}}
\newcommand{\vXi}{\vectg{\Xi}}
\newcommand{\vpsi}{\vectg{\psi}}
\newcommand{\vzeta}{\vectg{\zeta}}
\newcommand{\vzero}{\vect{0}}
\newcommand{\cA}{\mathcal{A}}
\newcommand{\cB}{\mathcal{B}}
\newcommand{\cC}{\mathcal{C}}
\newcommand{\cD}{\mathcal{D}}
\newcommand{\cE}{\mathcal{E}}
\newcommand{\cF}{\mathcal{F}}
\newcommand{\cG}{\mathcal{G}}
\newcommand{\cH}{\mathcal{H}}
\newcommand{\cI}{\mathcal{I}}
\newcommand{\cJ}{\mathcal{J}}
\newcommand{\cK}{\mathcal{K}}
\newcommand{\cL}{\mathcal{L}}
\newcommand{\cM}{\mathcal{M}}
\newcommand{\cN}{\mathcal{N}}
\newcommand{\cO}{\mathcal{O}}
\newcommand{\cP}{\mathcal{P}}
\newcommand{\cQ}{\mathcal{Q}}
\newcommand{\cR}{\mathcal{R}}
\newcommand{\cS}{\mathcal{S}}
\newcommand{\cT}{\mathcal{T}}
\newcommand{\cU}{\mathcal{U}}
\newcommand{\cV}{\mathcal{V}}
\newcommand{\cW}{\mathcal{W}}
\newcommand{\cX}{\mathcal{X}}
\newcommand{\cY}{\mathcal{Y}}
\newcommand{\cZ}{\mathcal{Z}}
\newcommand{\fA}{\mathfrak{A}}
\newcommand{\fB}{\mathfrak{B}}
\newcommand{\fC}{\mathfrak{C}}
\newcommand{\fD}{\mathfrak{D}}
\newcommand{\fE}{\mathfrak{E}}
\newcommand{\fF}{\mathfrak{F}}
\newcommand{\fG}{\mathfrak{G}}
\newcommand{\fH}{\mathfrak{H}}
\newcommand{\fI}{\mathfrak{I}}
\newcommand{\fJ}{\mathfrak{J}}
\newcommand{\fK}{\mathfrak{K}}
\newcommand{\fL}{\mathfrak{L}}
\newcommand{\fM}{\mathfrak{M}}
\newcommand{\fN}{\mathfrak{N}}
\newcommand{\fO}{\mathfrak{O}}
\newcommand{\fP}{\mathfrak{P}}
\newcommand{\fQ}{\mathfrak{Q}}
\newcommand{\fR}{\mathfrak{R}}
\newcommand{\fS}{\mathfrak{S}}
\newcommand{\fT}{\mathfrak{T}}
\newcommand{\fU}{\mathfrak{U}}
\newcommand{\fV}{\mathfrak{V}}
\newcommand{\fW}{\mathfrak{W}}
\newcommand{\fX}{\mathfrak{X}}
\newcommand{\fY}{\mathfrak{Y}}
\newcommand{\fZ}{\mathfrak{Z}}

\section{Additional Experiments}
    As described in the main submission, we complete our evaluation by incorporating the CIFAR100 as in-distribution setting as well as additional metrics.
    
    \subsection{CIFAR100 Results}
    To complete the AUROC evaluation discussed in the main paper, we find the corresponding results for the OOD detection setting with CIFAR100 as the in-distribution dataset in Table \ref{tab:stat:100} and Table \ref{tab:train:100}. 
    
    Table \ref{tab:stat:100} provides the OOD detection results for the \textbf{statistics} stream but does not include the Energy-based model~\cite{liu2020energy} due to no published results existing for the CIFAR100 setting. The results from Table \ref{tab:stat:100} reinforce the findings from the main submission, with the gap between the top performers (HDFF and Gram) growing significantly between the other methods, averaging out to $\approx$10\% AUROC difference. Comparative to these shifts in scores, the difference between HDFF and Gram remains small with HDFF taking significantly less computational time to attain its respective results. 
    
    Table \ref{tab:train:100} displays the additional OOD detection results for the \textbf{training} stream but does not contain NMD~\cite{Dong_2022_CVPR} due to the absence of published results in the CIFAR100 setting. Similarly to the results from the main submission, we see that HDFF in combination with other state-of-the-art OOD detectors increases performance across the majority of benchmarks. Specifically, we see that HDFF-1DS outperforms the Spectral Discrepancy Detector~\cite{Zaeemzadeh_2021_CVPR} in two of the four comparative benchmarks despite HDFF requiring 50x less computation to achieve these results. We note that in this CIFAR100 setting, Table \ref{tab:train:100} shows that HDFF-MLP is weaker at the SVHN and CIFAR10 OOD datasets. Due to the high compatibility of HDFF with the MLP on other ODO benchmarks, it is unclear if this drop is due to the MLP or the HDFF representation without comparative results from other representations. Future work on auxiliary OOD detection networks may investigate the sensitivity of these auxiliary networks to their input data, determining which representations are most effective for individual OOD datasets.
    
    \begin{table}[t]
        \centering
        \scriptsize
        \begin{tabular}{l|ccccc}
        \multicolumn{6}{c}{Statistics Stream - CIFAR100} \\ \hline
        \multicolumn{1}{c|}{OOD} & HDFF & HDFF-Ens & Gram & MSP & ML \\
        \multicolumn{1}{c|}{Dataset} & (Ours) & (Ours) & \cite{pmlr-v119-sastry20a} & \cite{hendrycks17baseline} & \cite{hendrycks2019anomalyseg} \\ \hline
            iSun & 95.2 & \second{95.8} & \best{98.8} & 82.5 & 85.5  \\ 
            TINc & 93.1 & \second{93.8} & \best{98.2} & 83.5 & 86.3  \\ 
            TINr & 95.4 & \second{96.0} & \best{98.5} & 81.6 & 84.3  \\ 
            LSUNc & 91.7 & \second{92.5} & \best{96.0} & 83.9 & 86.5  \\
            LSUNr & 94.5 & \second{95.3} & \best{99.3} & 82.7 & 85.5  \\ 
            SVHN & \second{99.2} & \best{99.4} & 99.0 & 86.7 & 90.0  \\ 
            MNIST & \second{99.8} & \second{99.8} & \best{99.9} & 82.4 & 84.6  \\ 
            KMNIST & 99.5 & \second{99.6} & \best{99.99} & 86.6 & 87.5  \\ 
            FMNIST & \second{98.4} & \second{98.4} & \best{99.4} & 91.0 & 93.3  \\ 
            DTD & 92.9 & \second{93.5} & \best{97.5} & 78.1 & 79.7  \\ 
            CIFAR10 & 65.7 & 68.2 & 74.2 & \second{80.9} & \best{81.5} \\ \hline
            \multicolumn{1}{c|}{\textbf{Average}}
              & 93.2 & \second{93.8} & \best{96.4} & 83.6 & 85.9  \\
        \end{tabular}
        \caption{OOD detection results for the against the methods contained belonging to the \textbf{statistics} stream with CIFAR100 as the in-distribution dataset. Comparison metric is AUROC, higher is better. \best{Best} results are shown in \best{blue and bold}, \second{second} best results are shown in \second{green and italics}. The ensemble in \textit{HDFF-Ens} always consists of 5 models. Due to an absence of published data, Energy-based model~\cite{liu2020energy} is not included. In the far-OOD settings, HDFF and Gram achieve significant ($\approx$10\%) improvements in AUROC over the competing methods.}
        \label{tab:stat:100}
    \end{table}
    
    \begin{table}[t]
        \centering
        \scriptsize
        \begin{tabular}{l|ccccc}
        \multicolumn{6}{c}{Training Stream - CIFAR100} \\ \hline
        \multicolumn{1}{c|}{OOD} & HDFF-MLP & HDFF-1DS & Spectral & DDU & MOOD \\
        \multicolumn{1}{c|}{Dataset} & (Ours) & (Ours) & \cite{Zaeemzadeh_2021_CVPR} & \cite{mukhoti2022deep} & \cite{Lin_2021_CVPR} \\ \hline
            iSun & \best{99.9} & \second{94.4} & - & - & 77.8\\ 
            TINc & \best{99.4} & \second{93.5} & 88.6 & 83.13* & - \\ 
            TINr & \best{99.8} & \second{94.0} & 93.7 & 83.13* & - \\ 
            LSUNc & \second{93.9} & 90.5 & 93.8 & - & \best{96.8}\\ 
            LSUNr & \best{99.96} & 94.9 & \second{95.7} & - & 77.6\\ 
            SVHN & 54.1 & \best{92.8} & - & \second{87.53} & 85.9\\ 
            MNIST & \best{99.8} & \second{96.9} & - & - & 91.3\\ 
            KMNIST & \best{99.6} & \second{98.4} & - & - & 97.2 \\ 
            FMNIST & \best{99.7} & 97.8 & - & - & \second{99.1} \\ 
            DTD & \best{91.0} & \second{86.4} & - & - & 71.7\\ 
            CIFAR10 & \second{44.9} & \best{77.7} & - & - & - \\ \hline
            \multicolumn{1}{c|}{\textbf{Average}}
             & 89.3 & \second{92.5} & \best{93.0} & 84.6 & 87.2 \\
        \end{tabular}
        \caption{OOD detection results for the against the methods contained belonging to the \textbf{training} stream with CIFAR100 as the in-distribution dataset. Comparison metric is AUROC, higher is better. \best{Best} results are shown in \best{blue and bold}, \second{second} best results are shown in \second{green and italics}. Due to an absence of published data, NMD~\cite{Dong_2022_CVPR} is not included. When HDFF is combined with other state-of-the-art detectors it consistently provides state-of-the-art performance across many benchmarks.}
        \label{tab:train:100}
    \end{table}

    \subsection{Additional Metrics}
        Tables \ref{tab:stat:FPR95} and \ref{tab:stat:DetErr} complete the \textbf{statistics} stream evaluation from our main submission against the current state-of-the-art under the FPR95 and Detection Error metrics respectively with both CIFAR in-distribution datasets. Note that neither Table \ref{tab:stat:FPR95} nor Table \ref{tab:stat:DetErr} contain results for the Energy-based Model~\cite{liu2020energy} due to published results being absent for these metrics. Across the board, Tables \ref{tab:stat:FPR95} and \ref{tab:stat:DetErr} reflect the results from the AUROC evaluation in the main submission with HDFF and Gram~\cite{pmlr-v119-sastry20a} outperforming the other comparison methods. Consistent with the results from the main submission, we see that the performance gap between HDFF and Gram is relatively minor, by comparison to the gaps of other methods, with HDFF having a significant lower computational cost than Gram. Overall, the original evaluation against the AUROC metric provides an accurate assessment of the relative performance of the compared methods, with Tables \ref{tab:stat:FPR95} and \ref{tab:stat:DetErr} reinforcing the findings from the original submission.
        
    \begin{table}[t]
        \centering
        \scriptsize
        \begin{tabular}{c|l|ccccc}
        \multicolumn{7}{c}{Statistics Stream - FPR95} \\ \hline
        \multicolumn{1}{c|}{ID} & \multicolumn{1}{c|}{OOD} & HDFF & HDFF-Ens & Gram & MSP & ML \\
        \multicolumn{1}{c|}{Dataset} & \multicolumn{1}{c|}{Dataset} & (Ours) & (Ours) & \cite{pmlr-v119-sastry20a} & \cite{hendrycks17baseline} & \cite{hendrycks2019anomalyseg} \\ \hline
           \multirow{11}[0]{*}{CIFAR10}
            & iSun & \second{2.7} & 2.8 & \best{0.6} & 21.8 & 10.5  \\ 
            & TINc & 6.8 & \second{6.6} & \best{2.7} & 28.4 & 16.2  \\ 
            & TINr & \second{2.7} & \second{2.7} & \best{1.0} & 29.9 & 17.0  \\ 
            & LSUNc & \second{20.7} & 20.8 & \best{8.8} & 25.7 & 14.3  \\ 
            & LSUNr & 2.1 & \second{1.8} & \best{0.4} & 21.4 & 10.3  \\ 
            & SVHN & 2.7 & \best{2.3} & \second{2.6} & 25.3 & 14.3  \\ 
            & MNIST & \second{0.01} & \best{0} & \best{0} & 56.6 & 43.9  \\ 
            & KMNIST & 1 & \second{0.4} & \best{0} & 42.4 & 31.4  \\ 
            & FMNIST & \second{1.2} & \best{0.3} & \best{0.3} & 37.2 & 25.1  \\ 
            & DTD & 27.3 & \second{26.4} & \best{8.7} & 46.7 & 39.3  \\ 
            & CIFAR100 & 83.6 & 82.8 & 68.1 & \second{52.8} & \best{47.8} \\ \hline
            \multirow{11}[0]{*}{CIFAR100}
            & iSun & 25.3 & \second{22.9} & \best{6.3} & 72.4 & 68.9  \\ 
            & TINc & 35.2 & \second{33.1} & \best{10.0} & 67.5 & 63.4  \\ 
            & TINr & 23.9 & \second{21.3} & \best{6.7} & 72.0 & 68.7  \\ 
            & LSUNc & 39.1 & \second{37.3} & \best{23.3} & 70.1 & 66.7  \\ 
            & LSUNr & 29.8 & \second{26.0} & \best{5.9} & 72.1 & 68.3  \\ 
            & SVHN & \second{3.1} & \best{2.4} & 4.1 & 66.1 & 59.2  \\ 
            & MNIST & \best{0} & \best{0} & \best{0} & 83.3 & \second{82.5}  \\ 
            & KMNIST & 0.04 & \second{0.01} & \best{0} & 70.2 & 70.3  \\ 
            & FMNIST & \second{6.7} & \second{6.7} & \best{1.7} & 49.0 & 40.9  \\ 
            & DTD & 31.9 & \second{30.5} & \best{15.1} & 79.3 & 78  \\ 
            & CIFAR10 & 92.2 & 91.0 & 86.8 & \second{77.2} & \best{76.4} \\ \hline
            \multicolumn{2}{c|}{\textbf{Average}}
             & 19.9 & \second{19.0} & \best{11.5} & 53.1 & 46.1 \\
        \end{tabular}
        
        \caption{OOD detection results for the against the methods contained belonging to the \textbf{statistics} stream. Comparison metric is FPR@95, lower is better. \best{Best} results are shown in \best{blue and bold}, \second{second} best results are shown in \second{green and italics}. Due to an absence of published data, Energy-based model~\cite{liu2020energy} is not included. HDFF and Gram consistently produce state-of-the-art level performance, with a significant margin between them and the other statistical baselines.}
        \label{tab:stat:FPR95}
    \end{table}
    
    \begin{table}[t]
        \centering
        \scriptsize
        \begin{tabular}{c|l|ccccc}
        \multicolumn{7}{c}{Statistics Stream - Detection Error} \\ \hline
        \multicolumn{1}{c|}{ID} & \multicolumn{1}{c|}{OOD} & HDFF & HDFF-Ens & Gram & MSP & ML \\
        \multicolumn{1}{c|}{Dataset} & \multicolumn{1}{c|}{Dataset} & (Ours) & (Ours) & \cite{pmlr-v119-sastry20a} & \cite{hendrycks17baseline} & \cite{hendrycks2019anomalyseg} \\ \hline
           \multirow{11}[0]{*}{CIFAR10}
            & iSun & \second{3.7} & 3.9 & \best{1.8} & 8.2 & 6.9  \\ 
            & TINc & 5.5 & \second{5.4} & \best{3.5} & 9.4 & 8.3  \\ 
            & TINr & \second{3.8} & \second{3.8} & \best{2.1} & 9.8 & 8.6  \\ 
            & LSUNc & 9.7 & 9.7 & \best{6.7} & 9.0 & \second{7.9}  \\ 
            & LSUNr & \second{3.3} & 3.4 & \best{1.2} & 8.0 & 6.7  \\ 
            & SVHN & 3.7 & \best{3.5} & \second{3.6} & 7.8 & 7.2  \\ 
            & MNIST & 1.5 & \second{1.1} & \best{0.1} & 16 & 15.4  \\ 
            & KMNIST & 2.9 & \second{2.6} & \best{0.3} & 12.3 & 12.0  \\ 
            & FMNIST & 3.1 & \second{2.5} & \best{1.6} & 11.9 & 10.7  \\ 
            & DTD & 12.8 & \second{12.7} & \best{6.7} & 15.3 & 15.6  \\ 
            & CIFAR100 & 29.9 & 30.0 & 28.0 & \best{17.4} & \second{18.0} \\ \hline
            \multirow{11}[0]{*}{CIFAR100}
            & iSun & 11.8 & \second{11.0} & \best{5.0} & 25.2 & 22.0  \\ 
            & TINc & 14.6 & \second{13.6} & \best{6.4} & 24.5 & 21.2  \\ 
            & TINr & 11.4 & \second{10.6} & \best{5.4} & 26.0 & 23.0  \\ 
            & LSUNc & 16.9 & \second{15.7} & \best{11.2} & 23.7 & 20.6  \\ 
            & LSUNr & 12.6 & \second{11.4} & \best{4.9} & 24.9 & 21.8  \\ 
            & SVHN & \second{3.9} & \best{3.5} & 4.3 & 20.7 & 16.7  \\ 
            & MNIST & 0.7 & \best{0.5} & \second{0.6} & 23.1 & 20.6  \\ 
            & KMNIST & \second{1.6} & \second{1.6} & \best{0.3} & 20.0 & 18.8  \\ 
            & FMNIST & 5.0 & \second{4.9} & \best{3.0} & 16.9 & 13.8  \\ 
            & DTD & 15.3 & \second{14.4} & \best{8.2} & 28.0 & 26.6  \\ 
            & CIFAR10 & 37.1 & 35.1 & 32.8 & \second{25.9} & \best{25.3} \\ \hline
            \multicolumn{2}{c|}{\textbf{Average}}
             & 9.6 & \second{9.1} & \best{6.3} & 17.5 & 15.8 \\
        \end{tabular}
       
        \caption{OOD detection results for the against the methods contained belonging to the \textbf{statistics} stream. Comparison metric is Detection Error, lower is better. \best{Best} results are shown in \best{blue and bold}, \second{second} best results are shown in \second{green and italics}. Due to an absence of published data, Energy-based model~\cite{liu2020energy} is not included.}
         \label{tab:stat:DetErr}
    \end{table}

    \begin{table}[t]
        \centering
        \scriptsize
        \begin{tabular}{c|l|ccccc}
        \multicolumn{7}{c}{Training Stream - FPR95} \\ \hline
        \multicolumn{1}{c|}{ID} & \multicolumn{1}{c|}{OOD} & HDFF & HDFF & Spectral & NMD & MOOD \\ 
        \multicolumn{1}{c|}{Dataset} & \multicolumn{1}{c|}{Dataset} & (MLP) & (1DS) & \cite{Zaeemzadeh_2021_CVPR} & \cite{Dong_2022_CVPR} & \cite{Lin_2021_CVPR} \\ \hline
           \multirow{11}[0]{*}{CIFAR10}
            & iSun & \best{0.01} & 0.5 & - & \second{0.3} & 38.8 \\ 
            & TINc & \best{0.4} & \second{1.7} & 9.0 & 3.9 & -  \\ 
            & TINr & \best{0.1} & \second{1.3} & 7.6 & - & -   \\ 
            & LSUNc & 9.6 & 3.3 & \best{2.8} & 6.1 & \second{3.2}  \\ 
            & LSUNr & \best{0} &\second{0.4} & 3.4 & - & 36.2   \\ 
            & SVHN & 66.4 & \second{5.2} & - & \best{2.3} & 17.2  \\ 
            & MNIST & \best{0.2} & 2.8 & - & - & \second{0.4}   \\ 
            & KMNIST & 5.0 & \second{2.5} & - & - & \best{0.3}  \\ 
            & FMNIST & \second{0.3} & 4.1 & - & - & \best{0.1}   \\ 
            & DTD & 18.9 & \second{14.8} & - & \best{6.0} & 56.0   \\ 
            & CIFAR100 & 85.8 & \second{42.6} & - & \best{36.2} & -  \\ \hline
            \multirow{11}[0]{*}{CIFAR100}
            & iSun & \best{0.2} & \second{23.6} & - & - & 81.5   \\ 
            & TINc & \best{2.4} & \second{28.8} & 41.7 & - & -   \\ 
            & TINr & \best{0.7} & \second{24.3} & 47.2 & - & -   \\ 
            & LSUNc & \second{25.1} & 48.2 & 50.2 & - & \best{17.0}  \\ 
            & LSUNr & \best{0.1} & \second{21.6} & 43.0 & - & 81.2  \\ 
            & SVHN & 96.0 & \best{26.0} & - & - & \second{63.7}  \\ 
            & MNIST & \best{0} & \second{11.7} & - & - & 57.7   \\ 
            & KMNIST & \best{0.4} & \second{6.7} & - & - & 16.6  \\ 
            & FMNIST & \best{0.01} & 8.3 & - & - & \second{4.6}  \\ 
            & DTD & \best{31.1} & \second{50.9} & - & - & 86.8  \\ 
            & CIFAR10 & \second{97.1} & \best{87.6} & - & - & -  \\ \hline
            \multicolumn{2}{c|}{\textbf{Average}}
             & 20.0 & \second{19.0} & 25.6 & \best{9.1} & 35.1 \\
        \end{tabular}
       
        \caption{OOD detection results for the against the methods contained belonging to the \textbf{training} stream. Comparison metric is FPR@95, lower is better. \best{Best} results are shown in \best{blue and bold}, \second{second} best results are shown in \second{green and italics}. Due to an absence of published data, DDU~\cite{mukhoti2022deep} is not included. Consistent with the findings from the main submission, HDFF when combined with recent state-of-the-art OOD detectors produces improved results, setting a new state-of-the-art.}
         \label{tab:train:FPR95}
    \end{table}

    \begin{table}[!ht]
        \centering
        \scriptsize
        \begin{tabular}{c|l|cccc}
        \multicolumn{6}{c}{Training Stream - Detection Error} \\ \hline
        \multicolumn{1}{c|}{ID} & \multicolumn{1}{c|}{OOD} & HDFF-MLP & HDFF-1DS & Spectral & NMD \\ 
        \multicolumn{1}{c|}{Dataset} & \multicolumn{1}{c|}{Dataset} & (Ours) & (Ours) & \cite{Zaeemzadeh_2021_CVPR} & \cite{Dong_2022_CVPR}  \\ \hline
           \multirow{11}[0]{*}{CIFAR10}
            & iSun & \best{0.5} & 1.9 & - & \second{1.4}  \\ 
            & TINc & \best{1.6} & \second{2.9} & 6.8 & 4.4  \\ 
            & TINr & \best{0.8} & \second{2.5} & 6.2 & -  \\ 
            & LSUNc & 6.6 & \second{4.0} & \best{3.7} & 5.5  \\ 
            & LSUNr & \best{0.3} & \second{1.8} & 3.8 & -  \\ 
            & SVHN & 22.3 & \second{4.9} & - & \best{3.4}  \\ 
            & MNIST & \best{2.1} & \second{3.8} & - & -  \\ 
            & KMNIST & \second{4.7} & \best{3.7} & - & -  \\ 
            & FMNIST & \best{2.1} & \second{4.5} & - & -  \\ 
            & DTD & 10.6 & \second{8.7} & - & \best{5.4}  \\ 
            & CIFAR100 & 35.2 & \best{15.6} & - & \second{16.6} \\ \hline
            \multirow{11}[0]{*}{CIFAR100}
            & iSun & \best{1.1} & \second{14.0} & - & -  \\ 
            & TINc & \best{3.3} & \second{15.2} & 18.9 & -  \\ 
            & TINr & \best{1.7} & 14.3 & \second{14.2} & -  \\ 
            & LSUNc & \best{13.2} & 18.0 & \second{13.9} & -  \\ 
            & LSUNr & \best{0.8} & 13.1 & \second{11.3} & -  \\ 
            & SVHN & \second{46.4} & \best{15.4} & - & -  \\ 
            & MNIST & \best{0.6} & \second{7.3} & - & -  \\ 
            & KMNIST & \best{2.0} & \second{4.9} & - & -  \\ 
            & FMNIST & \best{1.3} & \second{5.7} & - & -  \\ 
            & DTD & \best{16.4} & \second{23.6} & - & -  \\ 
            & CIFAR10 & \second{49.9} & \best{26.6} & - & - \\ \hline
            \multicolumn{2}{c|}{\textbf{Average}}
             & 10.2 & \second{9.7} & 9.9 & \best{6.1} \\
        \end{tabular}
        
        \caption{OOD detection results for the against the methods contained belonging to the \textbf{training} stream. Comparison metric is Detection Error, lower is better. \best{Best} results are shown in \best{blue and bold}, \second{second} best results are shown in \second{green and italics}. Due to an absence of published data, DDU~\cite{mukhoti2022deep} and MOOD~\cite{Lin_2021_CVPR} are not included.}
        \label{tab:train:DetErr}
    \end{table}

    The results for the \textbf{training} stream are reported in Tables \ref{tab:train:FPR95} and \ref{tab:train:DetErr} for the FPR95 and Detection Error metrics respectively. Note that neither table contains results for the DDU~\cite{mukhoti2022deep} OOD detector and Table \ref{tab:train:DetErr} additionally does not contain MOOD~\cite{Lin_2021_CVPR}; the absence of both is due to a lack of published results. The results from Tables \ref{tab:train:FPR95} and \ref{tab:train:DetErr} repeat the same patterns as both the main submission and the additional results described in Table \ref{tab:train:100}. In general, HDFF-MLP provides the strongest performance across the majority of the benchmarks, with HDFF-1DS also boasting large performance benefits over the original Spectral Discrepancy Detector~\cite{Zaeemzadeh_2021_CVPR} in the CIFAR10 setting, with smaller benefits in the CIFAR100 setting. In summary, the results from Tables \ref{tab:train:FPR95} and \ref{tab:train:DetErr} help strengthen the claims main in the main submission by broadening the standard of good performance.

\section{Ablations}
     \subsection{Projection Dimensionality} \label{sec:dimablate}
\begin{figure}[t]
            \begin{subfigure}{0.95\columnwidth}
                \centering \includegraphics[width=1.0\columnwidth]{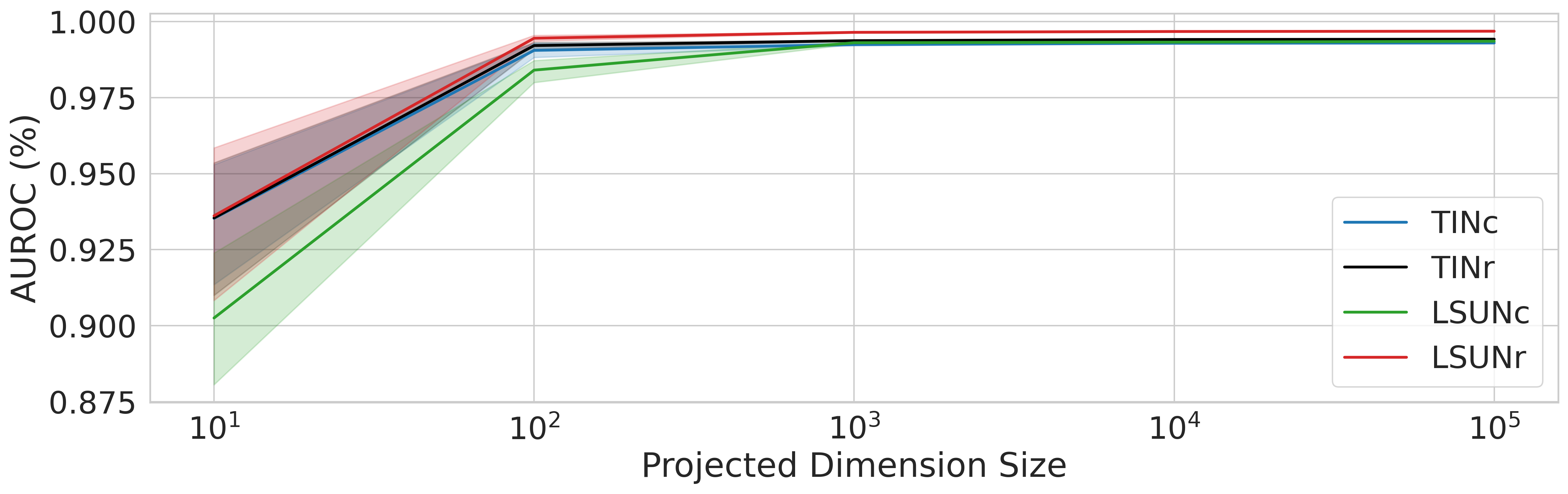}
                \caption{CIFAR10 as In-Distribution}
                \label{fig:1:DimAblate}
            \end{subfigure} \\
            \begin{subfigure}{0.95\columnwidth}
                \centering \includegraphics[width=1.0\columnwidth]{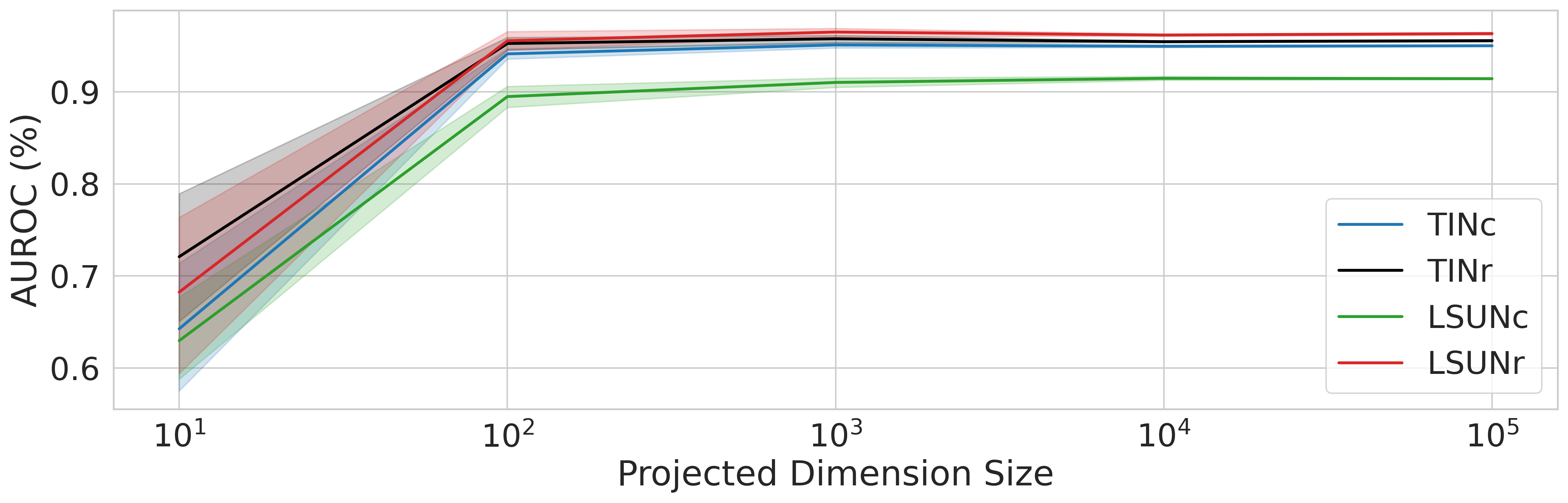}
                \caption{CIFAR100 as In-Distribution}
                \label{fig:2:DimAblate}
            \end{subfigure}
            \caption{Ablation of the required size of projected HD space to achieve best OOD detection performance. Individual plot lines show mean and 95\% confidence interval over 10 independent initialisations of projection matrices on the same model trained with the 1D Subspaces \cite{Zaeemzadeh_2021_CVPR} methodology. Features should be projected into a HD space in the range of $10^3 - 10^4$ dimensions to remove the potential for variation in OOD detection performance.
            }
            \label{fig:DimAblate}
        \end{figure}

        Figure \ref{fig:DimAblate} visualises the influence of the number of dimensions in the hyperdimensional space on the OOD detection performance.
        Specifically, we plot the mean and 95\% confidence interval over 10 initialisations of the random projection matrices at each order of magnitude for a single 1D Subspaces \cite{Zaeemzadeh_2021_CVPR} trained model. The general trend seen in both figures is that as the number of dimensions increases up to $10^3$, the mean performance of the HDFF detector increases. Similarly, we see that the bounds of 95\% confidence all but effectively disappear when considering spaces of $10^3$ dimensions or greater. The results of this ablation show that standard choices of hyperdimensional space in the range of $10^3 - 10^4$ will produce reasonable results, consistent with conventions used in HDC literature \cite{Neubert_2021,9388914,10.1145/3453688.3461749}. 

     \subsection{Additional Layer Ablations}
\begin{figure*}[t]
            \centering
            \begin{subfigure}{\scaleLayer}
                \centering
                \includegraphics[width=\columnwidth]{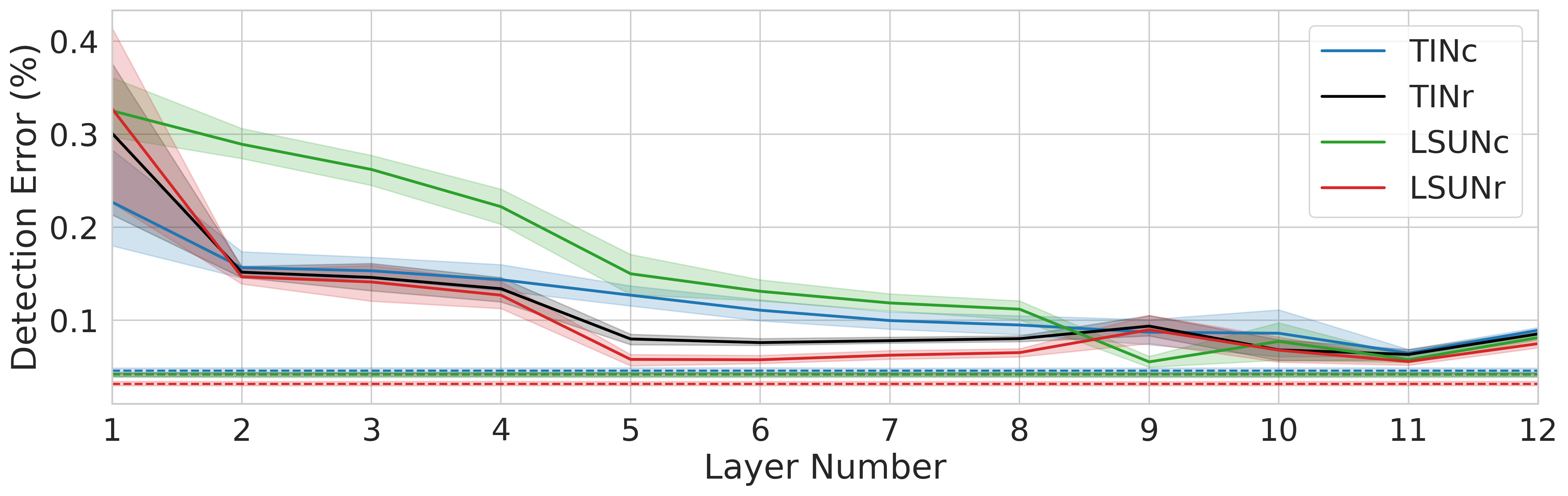}
                \caption{CIFAR10 - Detection Error}
                \label{fig:1:layer}
             \end{subfigure}
             \begin{subfigure}{\scaleLayer}
                \centering
                \includegraphics[width=\columnwidth]{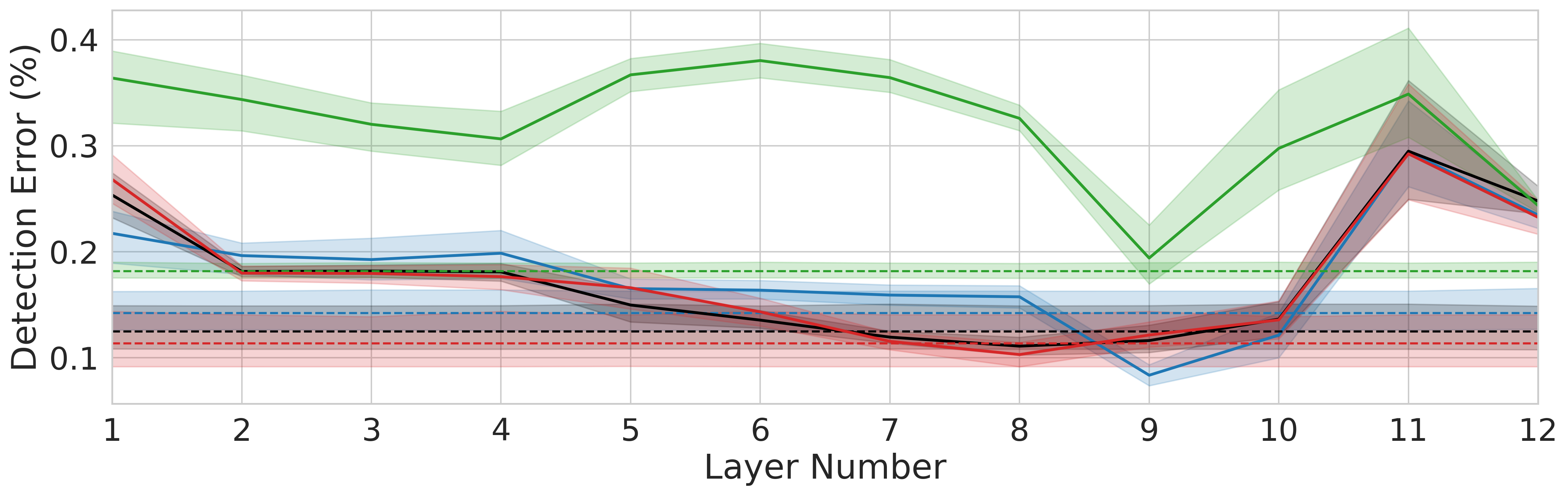}
                \caption{CIFAR100 - Detection Error}
                \label{fig:2:layer}
             \end{subfigure} \\
             \begin{subfigure}{\scaleLayer}
                \centering
                \includegraphics[width=\columnwidth]{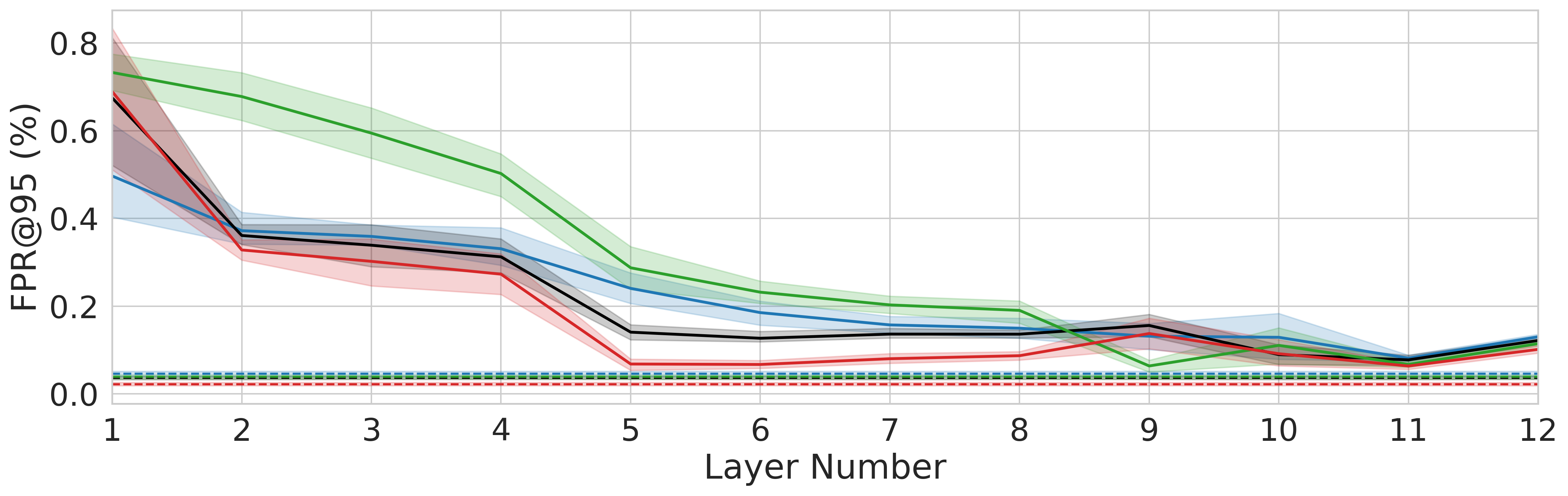}
                \caption{CIFAR10 - FPR@95}
                \label{fig:3:layer}
             \end{subfigure}
             \begin{subfigure}{\scaleLayer}
                \centering
                \includegraphics[width=\columnwidth]{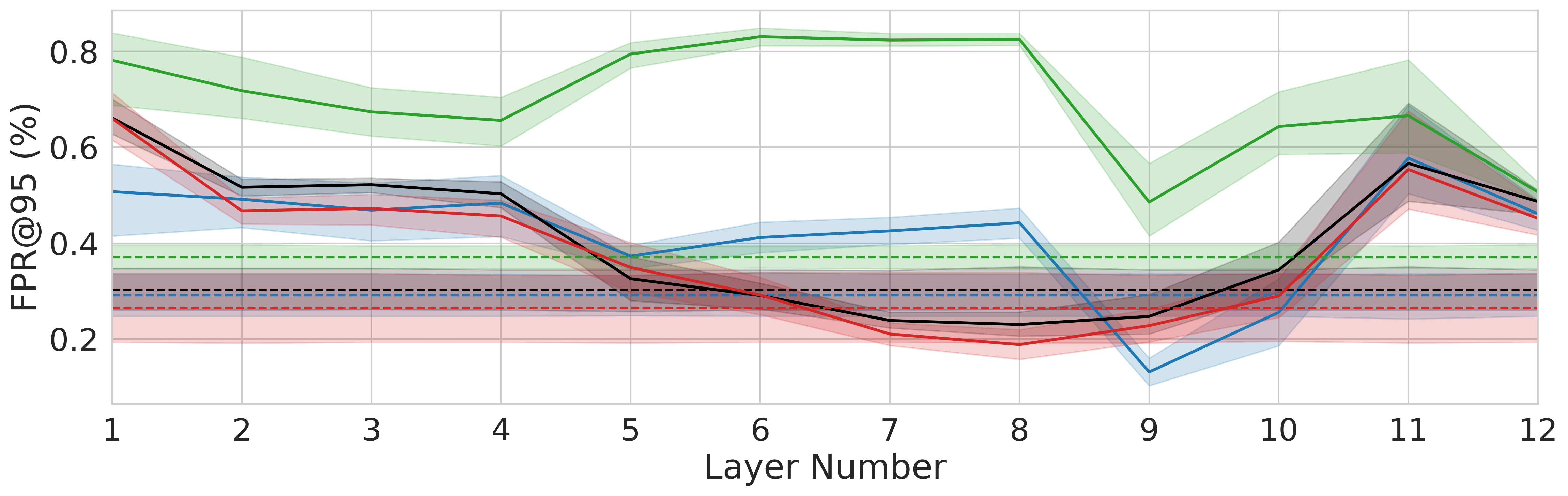}
                \caption{CIFAR100 - FPR@95}
                \label{fig:4:layer}
             \end{subfigure}\\
             \begin{subfigure}{\scaleLayer}
                \centering
                \includegraphics[width=\columnwidth]{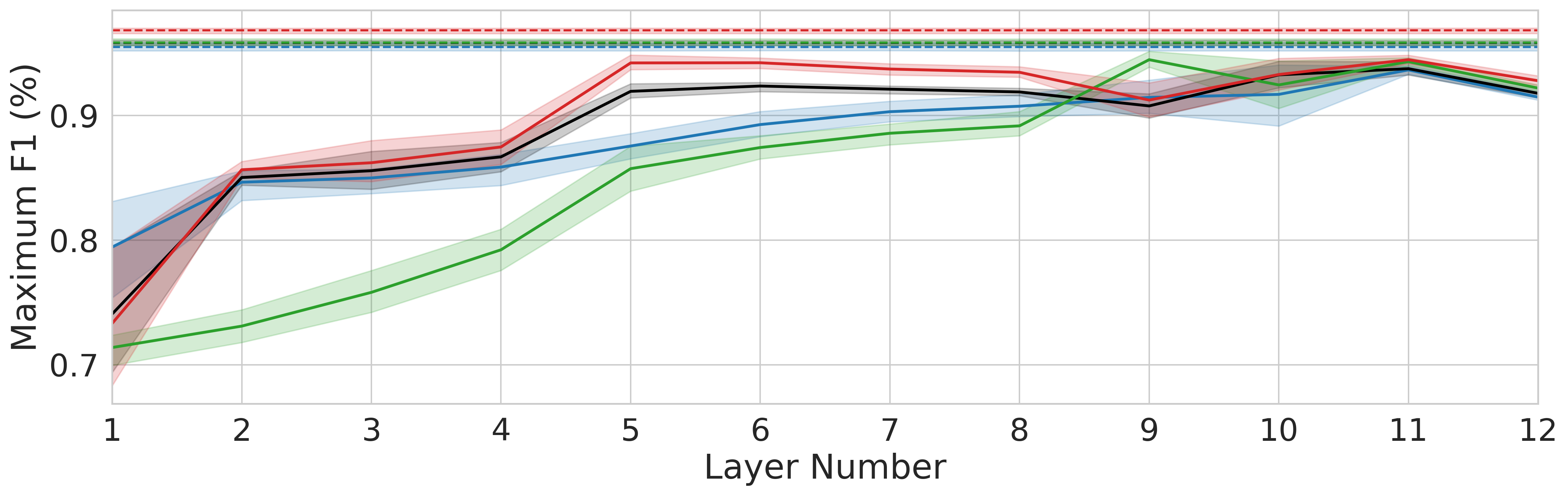}
                \caption{CIFAR10 - Maximum F1}
                \label{fig:5:layer}
             \end{subfigure}
             \begin{subfigure}{\scaleLayer}
                \centering
                \includegraphics[width=\columnwidth]{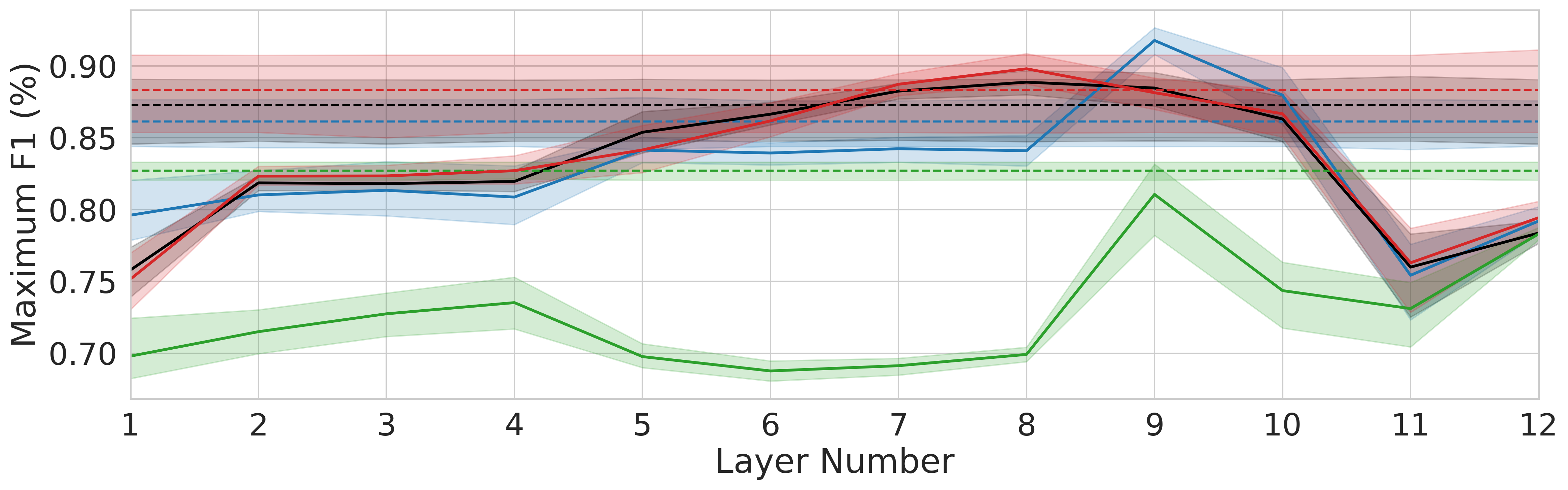}
                \caption{CIFAR100 - Maximum F1}
                \label{fig:6:layer}
             \end{subfigure}
             \caption{Layer-wise ablation across multiple metrics for the 1D Subspaces trained model. We plot mean and 95\% confidence interval for each datasets with the dotted lines corresponding to the performance of the fusion of feature maps across the network. For the CIFAR10 setting in (a), (c) and (e) it is clear that the fusion of feature maps from across the network provide the best performance. For the CIFAR100 setting in (b), (d) and (f) we see that violations of this characteristic are present in the later layers of 7, 8 and 9. Although some individual layers violate this characteristic, we note that no individual layer performs well across all of the OOD datasets.}
             \label{fig:layer}
        \end{figure*}

        Figure \ref{fig:layer} provides additional ablations of individual layer performance with respect to the other standard metrics for the benchmark defined in \cite{liang2018enhancing}; those being FPR@95, Detection Error and Maximum F1 Score. We make use of the same 12 BasicBlock hooks as in the main submission. Across both settings we see that the cropped datasets nearly perfectly mirror each other, indicating that there is a set of features that HDFF is consistently strong at detecting. In the CIFAR10 setting we see that across all metrics, there is no individual layer that performs at or above the level of the the fusion of feature maps. This trend is violated by the later layers, 7 through to 9 in the CIFAR100 setting. In particular, we see that the metrics that measure binarised performance, F1 and FPR@95, appear to have the largest increase above the fusion of features, usually in only a single layer. Overall, we make the same observation as in the main submission; there is no individual layer that performs well across all OOD datasets, assuming we already have the prior knowledge of how well each layer performs at this task. Future work into weighting individual layers based on predicted performance may be an important consideration. 
        
    \subsection{Preprocessing}
        In order to apply our hyperdimensional feature fusion to the feature maps of a DNN we first implemented a preprocessing pipeline that made use of a global pooling operation and mean-centering. Table \ref{tab:preprocess} ablates the choice of pooling components on a subset of datasets with the AUROC metric, showing their influence to the overall results described in the main paper.
        
        Table \ref{tab:preprocess} shows that in most cases max pooling increases performance in the CIFAR10 settings and average pooling provides better results on the whole for the CIFAR100 setting. We additionally note that the performance gaps between max and average pooling are relatively minor, with only the LSUN datasets in the CIFAR100 setting having a discrepancy greater than 1\%. Overall, both max and average pooling are approximately balanced with no clear better choice.
        
        \begin{table}[t]
            \centering
            \begin{tabular}{|l|l|l|l|}
            \hline
                ID set & OOD set & Avg Pool & Max Pool  \\ \hline
                \multirow{4}[0]{*}{CIFAR10} & TINc & 98.0 & \best{98.3}  \\ \
                 & TINr & 98.8 & \best{99.2}  \\ 
                 & LSUNc & \best{96.9} & 96.2  \\
                 & LSUNr & 99.1 & \best{99.2}  \\ \hline
                \multirow{4}[0]{*}{CIFAR100} & TINc & \best{93.7} & 93.1  \\ 
                 & TINr & \best{96.1} & 95.4  \\ 
                 & LSUNc & 87.7 & \best{91.7}  \\ 
                 & LSUNr & \best{95.8} & 94.5 \\ \hline
            \end{tabular}
            \caption{Ablation of our feature preprocessing in terms of the AUROC metric in the single-inference pass setting with the standard cross-entropy trained WideResNet model. \best{Best} result per ID + OOD dataset combination is coloured. Differing pooling operations provide better results dependant on the specific ID/OOD dataset configuration with no clear winner between the two.}
            \label{tab:preprocess}
        \end{table}
        
        \subsection{Inter-Sample Similarity}
\begin{figure*}[t]
            \centering
            \includegraphics[width=0.95\textwidth]{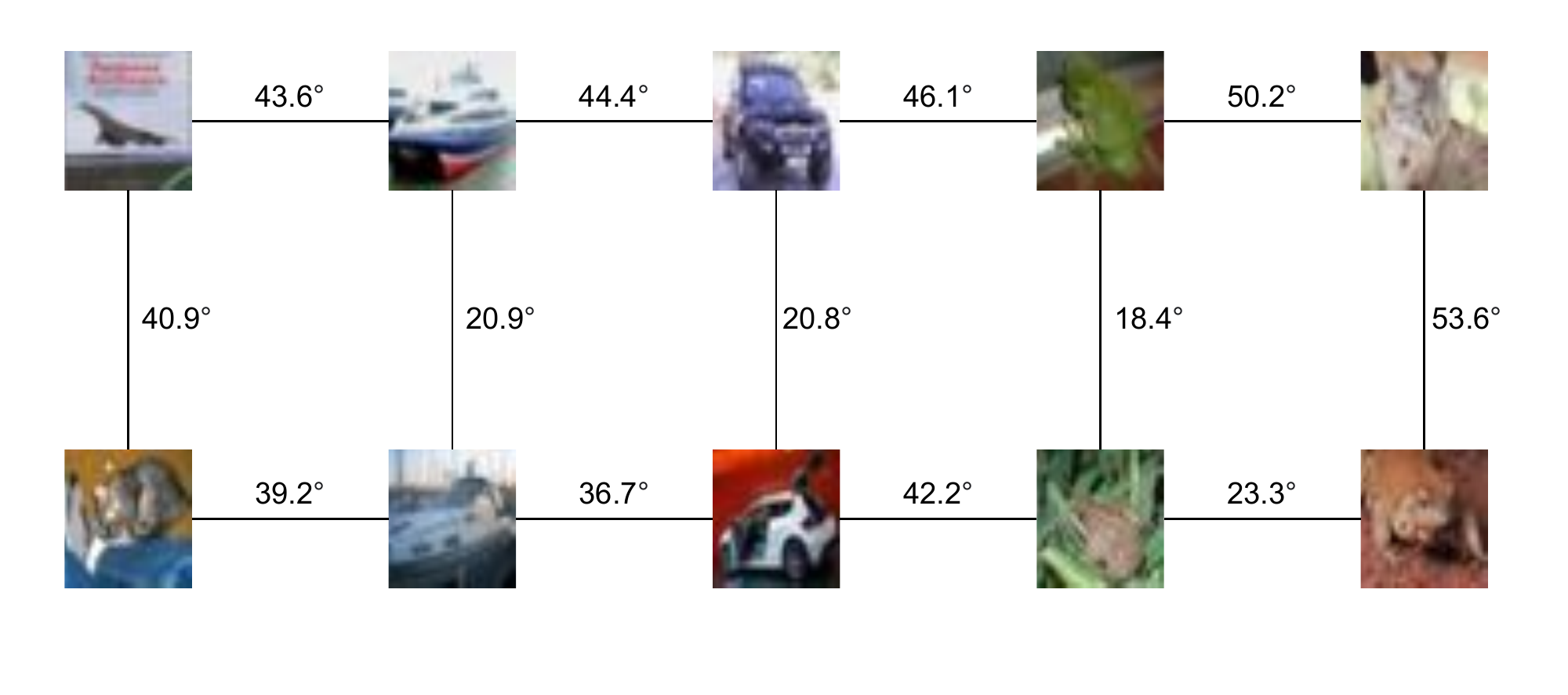}
            \caption{Visualisation of similarity between CIFAR10 ID samples using a 1D Subspaces trained WideResNet. Lines with angles between images corresponds to the angular distance between each images respective image descriptor vector. Comparisons are only made between direct neighbours to avoid clutter.}
            \label{fig:id_sim}
        \end{figure*}

        The concept of the angular distance between class descriptor vectors and image descriptors as visual similarity between the sample and the input extends to comparisons between pairs of image descriptor vectors. By comparing the descriptor vectors between two images, HDFF provides a quantitative metric that evaluates how visually similar the two input samples are. Since the angles between vectors is bounded between $[0, 90]$ degree we can assign semantic meaning to these bounds with 0 corresponding to exactly matching images and 90 corresponding to images with no similarity at all. Figure \ref{fig:id_sim} demonstrates a few examples of this, visualising the angular difference between images sampled from the CIFAR10 ID set.

{\small
\bibliographystyle{ieee_fullname}
\bibliography{refs}
}